\documentclass[11pt]{article}

\usepackage[final]{acl}

\usepackage{times}
\usepackage{latexsym}
\usepackage{booktabs}
\usepackage{amsmath}
\usepackage{amssymb}
\usepackage{float}
\usepackage{multicol}
\usepackage{tabularx}
\usepackage{graphicx}
\usepackage{subcaption}
\usepackage{booktabs}
\usepackage{multicol}
\usepackage{bm}
\usepackage{tabularx}
\usepackage{placeins}
\usepackage{float}

\usepackage[T1]{fontenc}

\usepackage[utf8]{inputenc}

\usepackage{microtype}

\usepackage{inconsolata}

\usepackage{graphicx}

\title{From Arraignment to Resolution: Transformer Models of Court Cases}

\author{Kory Rosen \and Aron Culotta\\
Department of Computer Science\\
Tulane University\\
New Orleans, LA 70118, USA\\
\texttt{\{krosen5, aculotta\}@tulane.edu}}

\begin{document}
\maketitle
\begin{abstract}
The criminal legal system is composed of many complex decision points, such as which charges are pursued, how bond is determined, plea deal negotiations, and sentencing. Given the profound impact of such decisions, there is a growing need to increase the transparency within this system. While prior machine learning approaches have focused on modeling single decisions (e.g., recidivism), this paper proposes a transformer-based time series model to forecast the outcome of the next proceeding in a case (e.g., charges accepted, bond set, plea deal made, etc.). By modeling a case as an event sequence, we enable a more nuanced analysis to support better oversight and understanding of decision making. We conduct an empirical analysis of thousands of cases using a newly constructed dataset derived from publicly available court records of a mid-sized U.S. city. We fit models that combine textual data from clerk notes with structured information about the defendant, prior charges, and courtroom. Our transformer model achieves $.57$ micro-F1 on a 20-class multi-label task, a significant improvement over several baselines and comparable to a GRU trained on the same input. We demonstrate how the model can be used to provide insights into decision-making and help court observers identify emerging issues within the jurisdiction studied. \footnote{Our code and synthetic dataset are available at \url{https://github.com/tapilab/docket_sequence_modeling}} 
\end{abstract}

\section{Introduction}

Each year, millions of people pass through criminal court in the United States. Each case is the product of a chain of decisions: initial appearance, bond determination, arraignment, continuances, plea negotiations, and, in a minority of cases, trial and sentencing. 
At every step, decisions are made by judges, prosecutors, and defense attorneys, and those decisions narrow the set of possible outcomes for the defendant. 
The cumulative effect is significant, determining whether a defendant spends weeks or months in pre-trial detention, what charges are pursued or dropped, and ultimately, what sentence is imposed. Yet, the patterns underlying these decisions are difficult to study at scale. Instead of structured event fields, what actually happens at each proceeding is often recorded in free-text clerk minutes. This combination of unstructured input and missing event labels is a practical barrier. Oversight bodies and researchers who would like to identify disparities or audit decision-making have few tools that operate at the level of the individual proceeding, because building one requires first converting clerk text into a consistent, machine-readable label of what occurred, and fitting proceeding-level outcome models.

Machine learning research on the criminal legal system has not, to date, addressed this gap. The dominant line of work focuses on a single, terminal prediction, such as recidivism, failure-to-appear, or pretrial release, and often uses structured features available at a single point in time \citep{angwin2016machinebias, doi:10.1126/sciadv.aao5580, Wang2023, 
10.1093/qje/qjx032, 10.1145/3442188.3445908}. This strain of research collapses an entire case into a single prediction target and obscures how decisions accumulate over the course of a case. No previous research provides a modeling framework for the question most relevant to our research: Given everything that has happened in a case so far, what is likely to happen next?

In this paper, we propose a transformer-based model that answers this question. We treat each case as a chronologically ordered sequence of proceedings, where each proceeding is represented by a multi-modal feature vector combining frozen pre-trained BERT embeddings of clerk notes with structured information about defendant demographics, charges, prior case history, bond amounts, procedural stage, and the assigned court section. At each step, our model predicts a multi-label vector over twenty outcome labels that span a full case, including bond decisions, charging decisions, probable cause determinations, plea entries, trial verdicts, and sentencing. Two features of the task require non-standard design choices. First, proceedings occur at irregular intervals rather than on a fixed temporal grid. To capture this, we represent absolute calendar time as a one-hot month-year vector concatenated into each proceeding's feature representation, in addition to standard learned positional embeddings that encode step order within a case. Second, multiple events frequently occur within a single proceeding, which we accommodate by applying multi-label supervision at each step.

We construct a balanced dataset of 2,533 cases and 19,459 proceedings from the publicly available court records of a mid-sized U.S. city. We evaluate our model under 5-fold cross-validation against three baselines: a frequency-based random classifier, a multi-layer perceptron that predicts final case outcomes from the features available at the first proceeding, and a GRU trained on the same proceeding representation as the transformer. The transformer model achieves a micro-F1 of $.5715$ on proceeding-level prediction, a $215\%$ improvement over the random baseline ($.1812$). The GRU reaches $.5886$, which is not statistically distinguishable from the transformer. When restricted to terminal outcomes for a fair comparison with the completion-only baseline, the transformer outperforms the MLP completion predictor by $.0831$ micro-F1. 

Our work offers four main contributions. First, we introduce the task of proceeding-level prediction as a new task for criminal justice analysis. Second, we propose a transformer architecture that handles irregular temporal spacing and heterogeneous multi-modal inputs, and evaluate it against baselines. Third, we construct a new dataset of proceeding-level court records annotated with twenty outcomes. The records themselves cannot be redistributed for legal and privacy reasons discussed in Appendix~\ref{sec:dataandcodeavail}. Instead, we release our full label extraction and modeling pipeline along with a schema-identical synthetic dataset. Fourth, we conduct a multi-method interpretability analysis that combines ablation studies, Integrated Gradients attributions, counterfactual sensitivity, and clustering of learned representations to characterize both what the model relies on and the structure it has discovered in the data. 

Beyond prediction, the model serves as an analytical tool for studying court decision-making in the jurisdiction studied. Counterfactual sensitivity analysis quantifies how predictions shift when case attributes are perturbed, showing that the model is most responsive to prior case count and less sensitive to race. Clustering the learned case representations reveals procedural trajectories, such as cases where charges are refused at the preliminary hearing stage, cases that proceed to trial, and detained defendants who plead guilty and receive jail time. Finally, gaps between confident predictions and observed outcomes flag potential procedural anomalies. Together, these applications let oversight bodies characterize what drives court decisions and flag cases within the studied jurisdiction for closer examination.

\section{Related Work}

Prior work on machine learning for  criminal court has focused almost exclusively on single-decision prediction. Risk assessments, such as COMPAS, have been scrutinized for cross-racial accuracy disparities \citep{angwin2016machinebias}. Algorithmic approaches to pretrial release have used gradient-boosted trees \citep{10.1093/qje/qjx032} and hierarchical Bayesian models \citep{10.1145/3442188.3445908} to study bail decisions. Legal Judgment Prediction (LJP) extends this paradigm to case text, using methods ranging from SVMs \citep{aletras2016predicting} and hierarchical BERT variants \citep{chalkidis-etal-2019-neural,niklaus-etal-2021-swiss} to joint multi-task frameworks that predict law articles, charges, and penalties simultaneously \citep{ZHANG2024112033,li-etal-2025-legal}. The work most structurally similar to ours is \citet{Polo_2021}, who classify Brazilian proceedings into a three-way status label from clerk-written motions using LSTMs and BERT; however, they collapse all proceedings into a single case-level label rather than predicting the next event. On the sequence modeling side, transformer models for health records \citep{Li2020BEHRT} and career trajectories \citep{vafa2024career} demonstrate that such architectures generalize well to irregular event streams with structured covariates, motivating our adoption of the same paradigm for criminal court. Our work differs from all of the above in that we predict the multi-label outcome of the next proceeding rather than a terminal case outcome. Alongside these computational approaches, courts are also studied through non-computational forms of oversight, such as court-watching programs. Court-watching programs recruit volunteers to observe proceedings and report on bail practices, presence of attorneys, and the conditions within the courtroom, and operate in many US jurisdictions, including Court Watch Massachusetts, Court Watch Chicago, Court Watch NOLA, and Court Watch New York City \citep{Gill2021-om, Reinke_Marx_2025}. These volunteers record what dockets omit, but their capacity limits the scope of their observations. A full survey of related work appears in Appendix~\ref{sec:related_work_full}.

\section{Dataset}

We construct our dataset from the criminal court records of a mid-sized U.S. city, where we have cases of varying complexity and duration. The criminal court system in this city is split across two court types by procedural stage, which we will refer to as Court A and Court B throughout the paper for the sake of anonymity. Court A handles proceedings in the early stages of a case, where events such as initial appearances and bail determinations would occur, while Court B deals with the proceedings of the later stages of the case, having to deal with arraignments, plea hearings, trials, and sentencing. We merge records from both to construct unified case-level sequences. 

\paragraph{Proceeding Information}
Each proceeding contains structured and unstructured information: (1) clerk notes which document judicial decisions, attorney requests, and procedural actions; (2) defendant demographics which include race (6 categories), gender (3 categories), and age at the initiation of a case; (3) criminal charges for an individual for a specific case which draws from a vocabulary of 2,139 charge types; (4) the month-year timestamp that corresponds to this proceeding; (5) bond information (original bond amounts, cumulative bond amounts); (6) the identification of Court A section (for proceedings that take place in Court A); (7) prior criminal history for a given defendant (prior case count and the aggregated prior charges) using only cases that precede the given proceeding.

\paragraph{Outcome Labels}
We extract $K = 20$ outcome labels representing key legal events from clerk notes using a regular expression pipeline (Table~\ref{tab:label_distribution}). 
These outcomes span the full length of a case, from initial hearings to final case disposition. A subset of 5 outcomes represents case-terminating events (charges refused, probable cause not found, trial verdict reached, sentences with or without jail time), which we evaluate separately in our baseline comparisons.

We developed the extraction patterns through iterative review of matched and unmatched clerk notes. For manual validation, we sampled and annotated 100 entries to ensure each outcome label appeared at least five times. We found no extraction errors in the sample.

\begin{table}[t]
\caption{Dataset statistics}
\label{tab:datasets}
\centering
\small
\begin{tabular}{@{}lr@{}}
\toprule
Cases & \textbf{2,533} \\
Total proceedings & \textbf{19,459} \\
Mean sequence length & \textbf{7.68} \\
Median sequence length & \textbf{7} \\
Max sequence length & \textbf{61} \\
Labels/proceeding & \textbf{1.63} \\
\bottomrule
\end{tabular}
\end{table}

\begin{table}[t]
\centering
\caption{Label distribution}
\label{tab:label_distribution}
\small
\begin{tabular}{lr}
\toprule
\textbf{Label} & \textbf{Count} \\
\midrule
Continuance Occurred & 5,424 \\
Bond Set & 3,703 \\
Court Appointed Public Defender & 3,173 \\
Pre-Trial Conference Held & 3,128 \\
First Appearance Hearing Occurred & 2,589 \\
Charges Refused & 1,848 \\
Defendant in Custody and Not in Court & 1,693 \\
Pled Not Guilty to Charges & 1,633 \\
Probable Cause Found & 1,431 \\
Pled Guilty to Charges & 1,077 \\
Bond Not Set & 991 \\
Defendant Released on Own Recognizance & 899 \\
Sentence Including Jail Time Imposed & 806 \\
State Was Not Present & 685 \\
Domestic Stay-Away Order Imposed & 546 \\
Charges Accepted & 462 \\
Ankle Monitoring Imposed & 456 \\
Probable Cause Not Found & 442 \\
Trial Verdict Reached & 404 \\
Sentence Without Jail Time Imposed & 401 \\
\midrule
\textbf{Total} & \textbf{31,791} \\
\bottomrule
\end{tabular}
\end{table}

\begin{figure}[t]
\centering
\includegraphics[width=\columnwidth, trim=0 0 0 .6cm, clip]{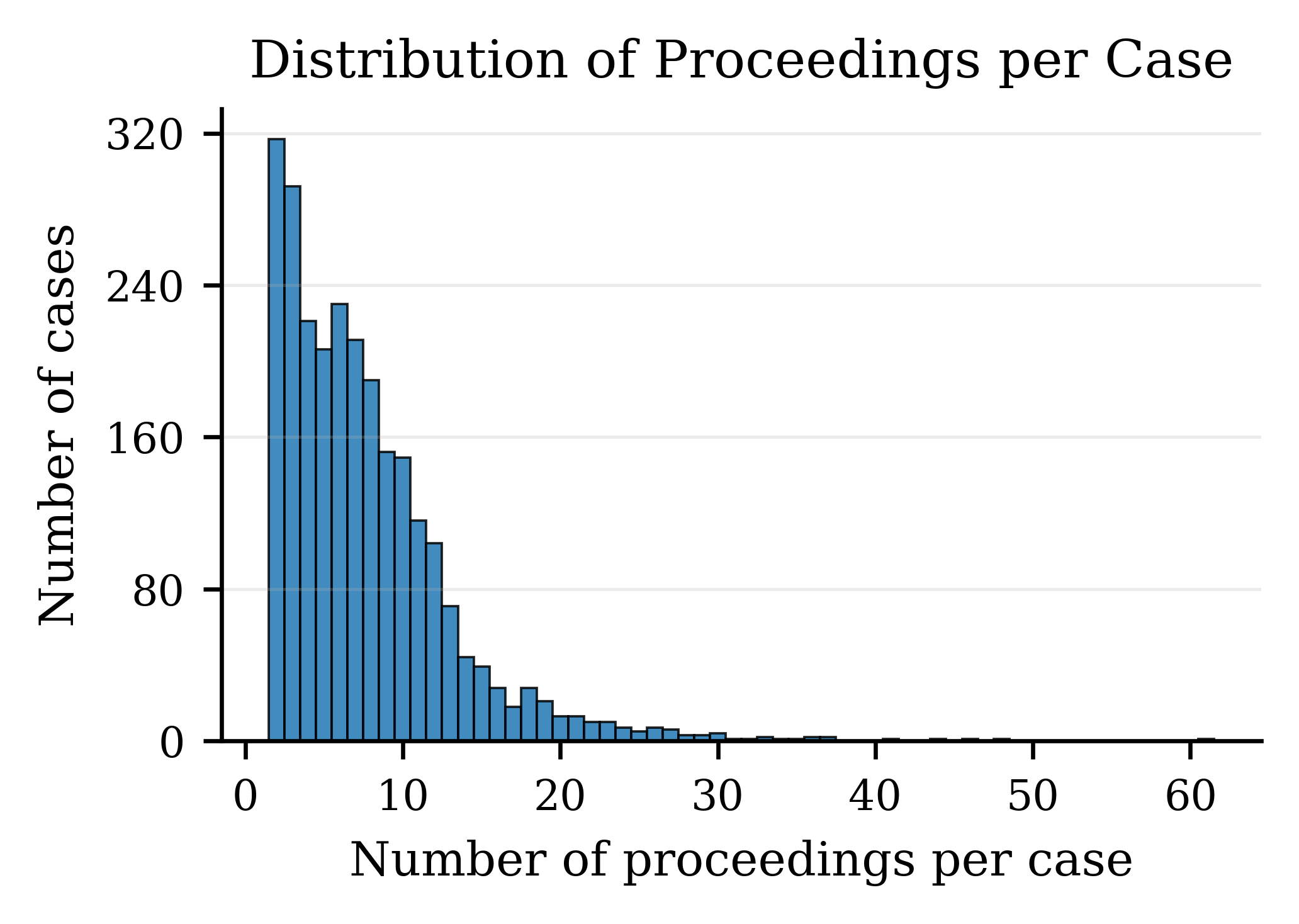}
\caption{Distribution of proceedings per case}
\label{fig:proceeding_distribution}
\end{figure}

\paragraph{Data Processing}
We apply several filtering steps, restricting to cases with complete defendant demographics, with a single defendant, with extractable outcome labels, and removing extradition cases that involve multiple jurisdictions. To address class imbalance, we employed stratified sampling to make sure that there is representation of rare but significant outcomes. This was done by first identifying outcomes that appear in fewer than 400 cases and including all such cases in our dataset, then randomly sampling additional cases to reach 2,533 cases.

Table~\ref{tab:datasets} presents key statistics of the balanced dataset. The dataset has major variations within case complexity, from sequences ranging from 2 to 61 proceedings and a median of 7 proceedings per case. Figure~\ref{fig:proceeding_distribution} shows the right-skewed nature of case lengths, with 58\% resolved in less than 8 proceedings, and a long tail up to 61 proceedings. On average, each proceeding has 1.6 labels, which reflects that these proceedings have multiple decisions made at a single proceeding.  Table~\ref{tab:label_distribution} shows the distribution of outcome labels across all proceedings in the dataset. The distribution shows significant class imbalance, as previously stated, and reflects the reality of criminal court proceedings, since most cases involve multiple procedural hearings before reaching a resolution, and most cases are resolved through plea deals.

\section{Methods}

\subsection{Notation and Setup}
Let $\mathcal{S} = \{(X^{(n)}, Y^{(n)})\}_{n=1}^{N}$ be our dataset of $N$ cases, where
\begin{itemize}
    \item $X^{(n)} = (p_0^{(n)}, p_1^{(n)}, \ldots, p_{T_n}^{(n)})$ is a chronologically ordered sequence of proceedings of variable length $T_n$;
    \item $Y^{(n)} = (y_1^{(n)}, y_2^{(n)}, \ldots, y_{T_n}^{(n)})$ is the corresponding sequence of multi-label outputs, with $y_t^{(n)} \in \{0,1\}^K$ a binary vector over $K = 20$ outcome labels.
\end{itemize}
At prediction step $t$, the model observes $p_0^{(n)}, \ldots, p_t^{(n)}$ and predicts the next outcome vector $y_{t+1}^{(n)}$, for $t = 0, 1, \ldots, T_n-1$.

\subsection{Proceeding Representation}

Each proceeding $p_t^{(n)}$ at time $t$ is represented by a fused feature vector that concatenates information about the case and defendant:
\begin{align*}
p_t^{(n)} ={} & f_{\text{fuse}}\big(
\psi_{\text{text}}(a_t^{(n)}),\;
\phi_{\text{demo}}(b^{(n)}), \\
& \phi_{\text{charges}}(c_t^{(n)}),\;
\phi_{\text{court\_a\_sec}}(d_t^{(n)}), \\
& \phi_{\text{prior\_labels}}(y_{t-1}^{(n)}),\;
\phi_{\text{charges}}(C_t^{(n)}), \\
& h_t^{(n)},\;
\psi_{\text{bond}}(j_t^{(n)}),\;
\psi_{\text{bond}}(k_t^{(n)}), \\
& l_t^{(n)},\;
\psi_{\text{time}}(\tau_t^{(n)})
\big)
\end{align*}
where $a_t^{(n)}$ is the proceeding text, $b^{(n)}$ encodes the defendant's demographics (race, gender, age at case initiation), $c_t^{(n)}$ is the charge-count vector of case $n$ as of proceeding $t$, and $d_t^{(n)}$ is the assigned section of Court A. The term $y_{t-1}^{(n)}$ is the multi-label vector at the previous proceeding, which is set to $\mathbf{0}^{K}$ at $t=1$. $C_t^{(n)}$ is the aggregated set of all prior charges for the defendant prior to the time of proceeding $p_t^{(n)}$. $h_t^{(n)} \in \{0, 1\}$ indicates whether the proceeding occurs in Court A or Court B. $j_t^{(n)}, k_t^{(n)}$ are the original and total bond amounts in dollars. $l_t^{(n)}$ is the prior case count of the defendant; $\tau(p_t^{(n)})$ is the month-year of the proceeding.

We write $\phi(\cdot)$ for learned embedding functions and $\psi(\cdot)$ for deterministic encoders. $\psi_{\text{text}}$ is a pretrained \texttt{bert-base-uncased} model \cite{devlin-etal-2019-bert} whose weights are frozen. $\psi_{\text{bond}}$ discretizes a bond amount into one of nine bins (including a ``not set / missing'' bin) and returns a one-hot vector. $\psi_{\text{time}}$ returns a one-hot month-year encoding. Scalars and binary indicators ($h_t^{(n)}$, $l_t^{(n)}$) are concatenated directly. (See Appendix~\ref{sec:embedding} for more details.)

\subsection{Model Architecture}
\label{sec:model_arch}

Given $p_{0:t}^{(n)}$, the model computes:
$$
\begin{aligned}
z_s^{(n)} &= p_s^{(n)} + \mathrm{PE}_s, \quad s = 0, 1, \dots, t, \\
u_{0:t}^{(n)} &= \mathrm{TransformerEncoder}\big( z_{0:t}^{(n)};\, M_{\text{causal}} \big), \\
\hat{y}_{t+1}^{(n)} &= \sigma\!\big( W_o\, u_t^{(n)} + b_o \big),
\end{aligned}
$$
where $\mathrm{PE} \in \mathbb{R}^{L_{\max} \times d}$ is the learned positional embedding, $M_{\text{causal}}$ is the upper-triangular causal mask, $W_o \in \mathbb{R}^{K \times d}$, $b_o \in \mathbb{R}^{K}$, and $\sigma$ is the element-wise sigmoid producing $K$ independent label probabilities. Under teacher forcing, the encoder output at position $t$ produces the prediction $\hat{y}_{t+1}^{(n)}$ for the labels of the next proceeding. At test time, the model never observes the fields from proceeding $t+1$ when predicting $y_{t+1}^{(n)}$.
The initial state of the sequence  omits all proceeding text, bonds, and prior label features, representing the information available prior to the first appearance hearing.

\subsection{Training objective}

We train the model by minimizing the average binary cross-entropy loss applied independently to each of the $K$ labels:
\begin{equation*}
\begin{aligned}
\mathcal{L}(\theta)
= -\frac{1}{N} \sum_{n=1}^{N} \frac{1}{T_n \cdot K}
\sum_{t=0}^{T_n-1} \sum_{k=1}^{K}\\
\Big[
y_{t+1,k}^{(n)} \log\left(\hat{y}_{t+1,k}^{(n)}\right)
+\\
\left(1-y_{t+1,k}^{(n)}\right)
\log\left(1-\hat{y}_{t+1,k}^{(n)}\right)
\Big].
\end{aligned}
\end{equation*}

\section{Experiments}

We compare with three baselines:

\paragraph{MLP Baseline} This baseline trains a static multi-layer perceptron that predicts the final case outcomes based  solely on the features available at the initial proceeding in order to determine if time sequence modeling is truly necessary for this task. The input includes the defendant demographics, initial charges, and Court A section, processed through a 3-layer network. 

\paragraph{Frequency-Based Random Classifier} This baseline establishes a lower bound on performance by sampling labels based on their training frequencies and ignoring the input features entirely. For each test proceeding, given its ground truth label count $k_t$, we sample $k_t$ labels without replacement. This evaluates our transformer's ability to perform better than a random guess, which has only been informed by the prevalence of a class in our training data.

\paragraph{GRU Baseline} This baseline trains a gated recurrent unit (GRU) on the same proceeding sequences as the transformer. For each proceeding $p_t^{(n)}$, the GRU updates its hidden state as $r_t^{(n)} = \mathrm{GRU}\big(p_{t}^{(n)};\, r_{t-1}^{(n)}\big)$ and predicts $\hat{y}_{t+1}^{(n)} = \sigma\!\big( W_o^{\mathrm{GRU}}r_t^{(n)} + b_o^{\mathrm{GRU}} \big)$. The two models share the same component embeddings, fusion layer, training objective, 896-dimensional hidden representations, optimizer, learning rate scheduler, and cross validation folds. Since recurrence is causal by design, the learned positional embedding $\mathrm{PE}$ and the causal mask $M_{\text{causal}}$ are not used for this baseline. We select the depth of the GRU based upon validation performance, which is 2 layers. 

We primarily evaluate model performance using the micro-averaged precision, recall, and F1 scores. The step-level results use top-$k_t$ selection, where $k_t$ is the number of outcomes actually observed at each proceeding. Therefore, precision, recall, and micro-F1 score are all equal at this level. We adopt this selection rule to avoid making our main results dependent on a classification threshold, while recognizing that it inflates absolute performance because $k_t$ would not be known in practice. Appendix~\ref{sec:threshold} reports the threshold-based results.  To establish a fair comparison with our first baseline model, we additionally report results in which the transformer's output is restricted to the subset of completion labels and evaluated only at the final input proceeding of each case. For additional implementation details and results, please see the Appendix.

\section{Classification Results}

We evaluate our model across three levels: step-level (prediction at each individual proceeding), completion-level (final case outcome prediction given full case history), and case-level (step-level predictions aggregated across the full case). Because these levels differ in what is predicted and when, metrics are not comparable across them. Step-level metrics are computed over individual proceedings, so a label recurring within a case contributes once per occurrence. On the other hand, case-level metrics are computed over cases, so each label contributes at most once per case. Therefore, support totals will differ. We report results at each level, followed by ablation, interpretability, and fairness analyses.

\newcolumntype{C}{>{\centering\arraybackslash}X}
\begin{table}[t]
\centering
\small
\setlength{\tabcolsep}{2pt}
\caption{Step-level performance (mean $\pm$ std)}
\label{tab:model_performance}
\begin{tabularx}{\columnwidth}{@{}lCCC@{}}
\toprule
Model & F1 & Precision & Recall\\
\midrule
Transformer
& $.5715 \pm .013$
& $.5715 \pm .013$
& $.5715 \pm .013$ \\
\addlinespace
GRU Baseline
& $.5886 \pm .013$
& $.5886 \pm .013$ 
& $.5886 \pm .013$\\
\addlinespace
Random Baseline
& $.1812 \pm .004$
& $.1812 \pm .004$
& $.1812 \pm .004$\\
\bottomrule
\end{tabularx}
\end{table}

\begin{table}[t]
\centering
\small
\setlength{\tabcolsep}{2pt}
\caption{Completion-level performance (mean $\pm$ std). Not comparable to Table~\ref{tab:model_performance}.}
\label{tab:completion_only_performance}
\begin{tabularx}{\columnwidth}{@{}>{\raggedright\arraybackslash}XCCC@{}}
\toprule
Model & F1 & Precision & Recall \\
\midrule
MLP Baseline
& $.4864 \pm .025$
& $.5118 \pm .027$
& \bm{$.4634 \pm .023$} \\
Transformer
& \bm{$.5695 \pm .014$}
& \bm{$.7534 \pm .010$}
& $.4579 \pm .016$ \\
\bottomrule
\end{tabularx}
\end{table}

Table~\ref{tab:model_performance} illustrates the step-level performance under 5-fold cross-validation, where the model predicts outcome labels after each individual proceeding. Our full transformer model achieves a micro-F1 score of $.5715 \pm .013$, which substantially outperforms the random baseline ($.1812 \pm .004$). This demonstrates that the model learns meaningful sequential patterns from the data. The GRU baseline reaches $.5886 \pm .013$, within fold-level variance of the transformer. Although the GRU performs slightly better, we use the transformer for the analyses done because its per-position outputs directly incorporate all prior proceedings rather than relying on a recursively compressed summary, making attributions and case representations comparable across positions within a case. 

Table~\ref{tab:completion_only_performance} reports the performance for completion-level models, where models observe the full case history and predict only the final case outcome. This task is quite different from the one shown above and should not be compared across tables. The transformer completion achieves a micro-F1 of $.5695 \pm .014$ which outperforms the MLP completion baseline ($.4864 \pm .025$) by $.0831$. This difference is statistically significant across the five folds ($95\%\text{ CI }[.053, .113];$ $t(4) = 7.76$, $p = .0015$).

Table~\ref{tab:per_label_transformer_3} reports the proceeding-level per-label performance for the full transformer under 5-fold cross-validation. The model performs strongest on frequent outcomes such as ``First Appearance Hearing Occurred'' and ``Court Appointed Public Defender'', but does poorest on rare outcomes such as ``Charges Accepted'', ``Probable Cause Not Found'', and ``Bond Not Set''. (See the Appendix for case-level results, where aggregation across proceedings raises performance substantially.)

\begin{table}[t]
\small
\caption{Proceeding-level per-label F1 performance (mean $\pm$ std over 5 folds), sorted by F1 score.}
\label{tab:per_label_transformer_3}
\begin{tabular}{lc}
\toprule
Label & F1 \\
\midrule
First Appearance Hearing Occurred & $.9926 \pm .00$ \\
Court Appointed Public Defender & $.7972 \pm .01$ \\
Bond Set & $.7351 \pm .01$ \\
Pre-Trial Conference Held & $.6576 \pm .02$ \\
Domestic Stay-Away Order Imposed & $.6479 \pm .02$ \\
Pled Not Guilty to Charges & $.6469 \pm .03$ \\
Charges Refused & $.4939 \pm .01$ \\
State Was Not Present & $.4914 \pm .06$ \\
Probable Cause Found & $.4743 \pm .02$ \\
Defendant Released on Own Recognizance & $.4549 \pm .05$ \\
Continuance Occurred & $.4426 \pm .02$ \\
Ankle Monitoring Imposed & $.4376 \pm .06$ \\
Sentence Without Jail Time Imposed & $.3793 \pm .04$ \\
Pled Guilty to Charges & $.3733 \pm .03$ \\
Sentence Including Jail Time Imposed & $.3388 \pm .04$ \\
Defendant in Custody and Not in Court & $.3075 \pm .03$ \\
Trial Verdict Reached & $.2838 \pm .04$ \\
Probable Cause Not Found & $.2349 \pm .04$ \\
Bond Not Set & $.1661 \pm .01$ \\
Charges Accepted & $.1340 \pm .02$ \\
\bottomrule
\end{tabular}
\end{table}

\paragraph{Ablation Studies}
\label{sec:ablation-studies}
To assess the contribution of individual input components, we conducted ablation experiments, systematically removing proceeding text, defendant demographics, prior labels, and combinations of these. Table~\ref{tab:ablation_results} shows that removing proceeding text yields the largest single-component drop (Micro-F1 of $.5335 \pm .016$, $\Delta \text{Micro-F1}$ of $-.0380$), which confirms that proceeding text is the primary driver of predictive performance. The combined removal of proceeding text and prior labels produces the largest degradation overall (Micro-F1 of $.5096 \pm .013$, $\Delta \text{Micro-F1}$ of $-.0619$), which indicates that these two components are jointly critical and partially complementary.

\begin{table}[t]
\centering
\small
\setlength{\tabcolsep}{3pt}
\caption{Ablation results under 5-fold cross validation relative to the transformer model (mean $\pm$ std)}
\label{tab:ablation_results}
\begin{tabularx}{\columnwidth}{@{}lCC@{}}
\toprule
Model & Micro F1 & $\Delta$ Micro-F1 \\
\midrule
Transformer
& $.5715 \pm .013$
& \\
No Text 
& $.5335 \pm .016$
& $-.0380$ \\
No Demographics 
& $.5682 \pm .014$
& $-.0033$ \\
No Prior Labels 
& $.5689 \pm .019$ 
& $-.0026$ \\
No Text or Prior Labels 
& $.5096 \pm .013$
& $-.0619$ \\
\bottomrule
\end{tabularx}
\end{table}

\section{Applying the model to court analysis}

Given the validated predictive model, we next turn to an examination into the ways in which such a model may be analyzed to provide insights into court decision making and to help court observers identify emerging issues of interest. The findings below reflect practices within the court studied. 

\subsection{Sensitivity}
\label{conterfactual-studies}

\begin{figure}[t]
    \centering
    \includegraphics[width=1\linewidth]{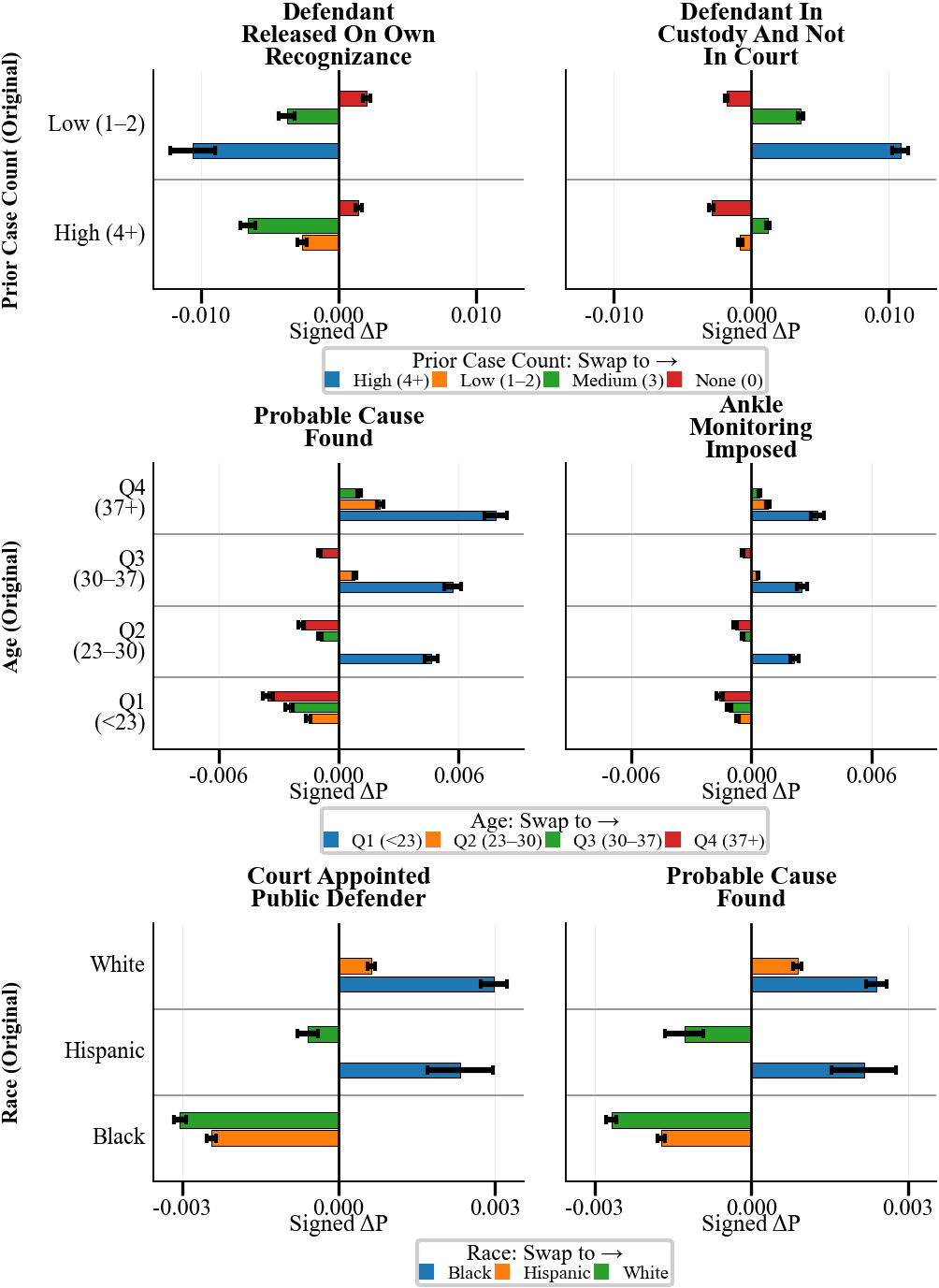}
    \caption{Counterfactual sensitivity analysis.}
    \label{fig:counterfactuals}
\end{figure}

First, we assess how sensitive the model is to demographic and case attributes. As the model has been trained on historical decision making, probing the model in this way provides a method to understand which factors may play a role in decision making. To do so, we conduct a counterfactual perturbation analysis~\cite{kusner2017}. For each prediction, we replace a feature value with an alternative value, holding other inputs fixed (e.g., the age of a defendant is modified from young to old). We then record the shift in posterior probability over the predicted labels that results from the feature substitution. We compute the average shift across all proceedings in the held out set (denoted $\Delta P$) and report 95\% confidence intervals using the $t$-distribution. Figure~\ref{fig:counterfactuals} shows sample per-label probability shifts for three features: prior case count, age, and race.

Overall, sensitivity is highest for legally relevant covariates and lowest for protected demographic attributes. Prior Case Count produces the largest perturbations, with a mean $\Delta P$ of $.0032$. The model predicts that defendants with high prior case counts are less likely to be released on their own recognizance (without paying bail) and are more likely to be in custody and not present at a hearing. These results reflect that prior court involvement is strongly associated with the court decisions recorded here, and also indicate that cases involving defendants with prior charges may have increased delays due to defendants missing hearings while in jail.

Age produces moderate perturbations, with a mean $\Delta P$ of $.0017$. The model predicts that younger defendants are more likely to have ankle monitoring imposed (electronic tracking device), to have a public defender appointed, and to have cases where probable cause is found. The increased probability for ankle monitoring may reflect increased perceived flight risk due to weaker community ties, or may reflect assumptions that younger defendants need greater oversight.

\begin{figure}[t]
    \includegraphics[width=1\linewidth, trim=0 0 0 1cm, clip]{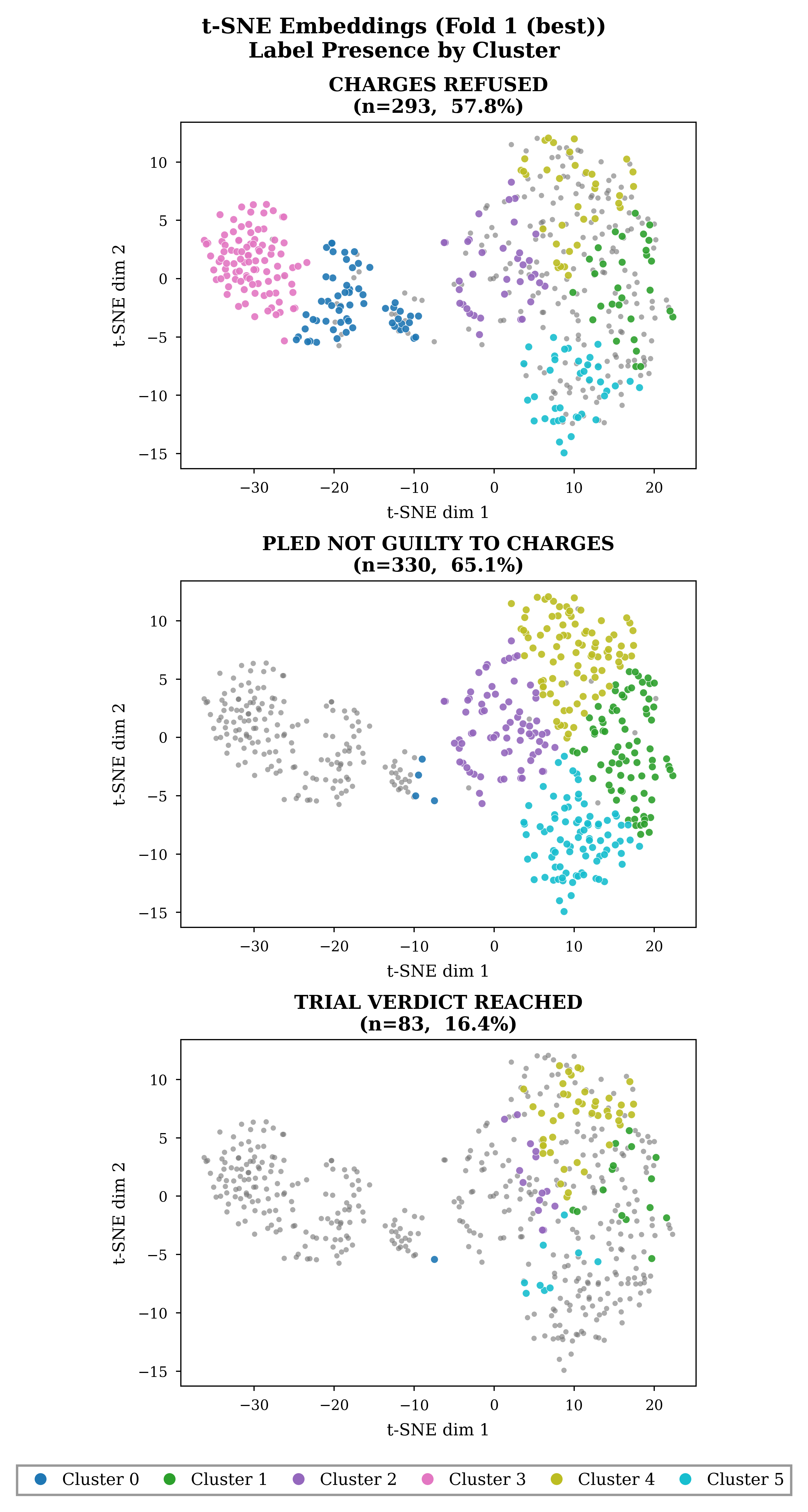}
    \caption{Clustered cases highlighting those containing three outcome labels.}
    \label{fig:tsnegrid}
\end{figure}

\begin{figure}[th!]
    \centering
    \includegraphics[width=1\linewidth]{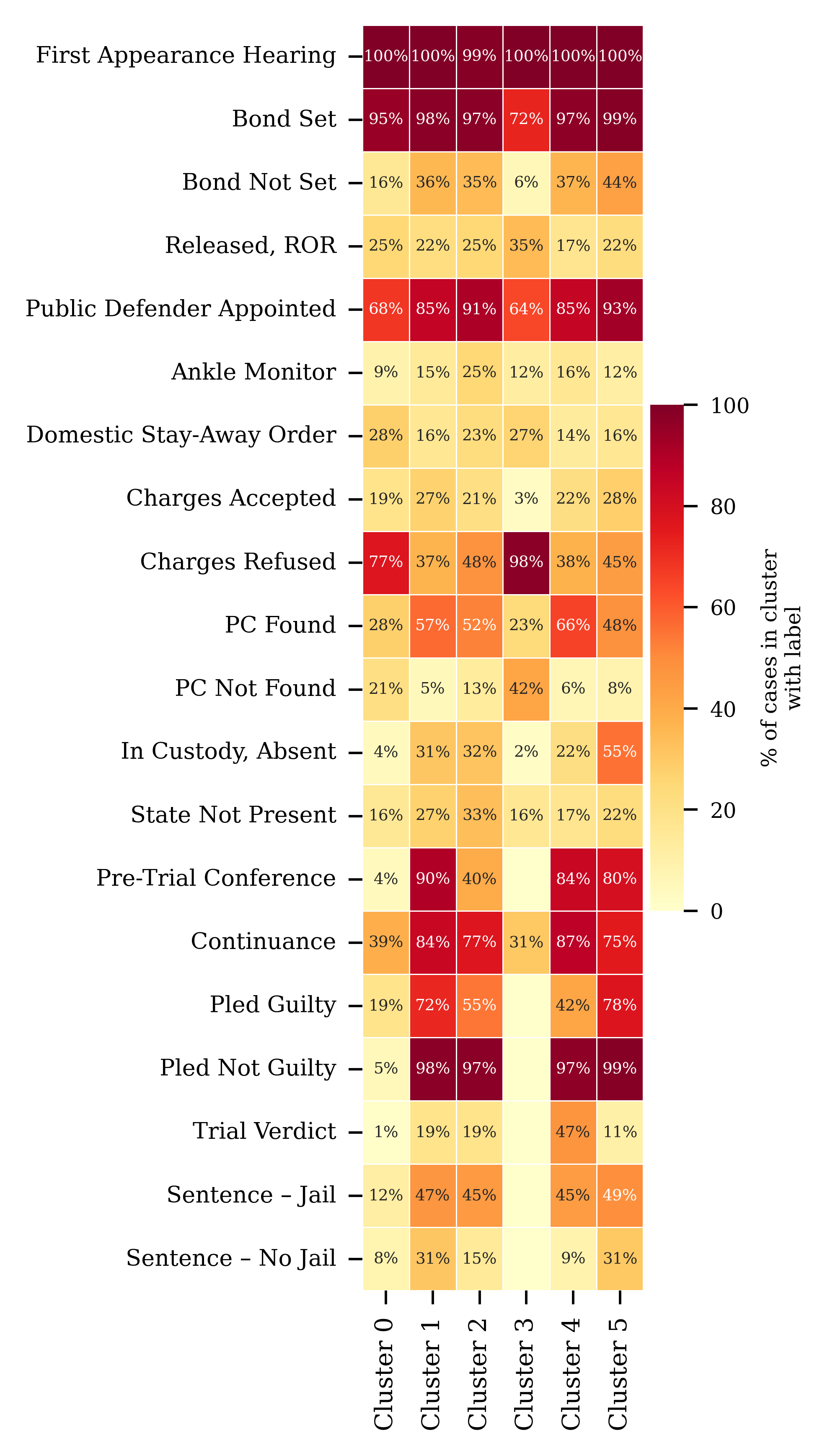}
    \caption{Labels by cluster.}
    \label{fig:tsne_heatmap}
\end{figure}

Race produces smaller perturbations, with a mean $\Delta P$ of $.0009$. We focus on the three largest groups in the dataset: Black, White, and Hispanic defendants. The model predicts that Black defendants are more likely to have a public defender appointed and more likely to have cases where probable cause is found. While the magnitudes are lower than for Prior Case Count and Age, the directionality is consistent with prior work on racial disparities in criminal courts~\cite{QuattroneCenter2017}.

\subsection{Clustering}

To discover groups of cases that exhibit similar temporal patterns, we perform a cluster analysis, using case representations learned by the transformer model. We extract the pre-classifier head transformer embeddings, apply StandardScaler normalization followed by L2 normalization, reduce to 50 principal components with Principal Component Analysis, and project to two dimensions with t-SNE. Then, we assign each case to a cluster by running K-Means ($K = 6$, $n_{\text{init}} = 10$) directly on the two-dimensional t-SNE coordinates.\footnote{We selected $K = 6$ after inspecting the elbow and silhouette curves for $ K = 2$ to 12.}

Figure~\ref{fig:tsnegrid} shows the per-label presence across the embedding space colored by cluster; Figure~\ref{fig:tsne_heatmap} shows the multi-label composition heatmap per cluster. The arrangement of cases in the t-SNE projection corresponds roughly to differences in their procedural trajectories. Variation across Dimension 1 appears to be associated with case time and whether charges are accepted. Dimension 2 is more nuanced, though cases with higher values appear to correspond to cases that conclude with a trial verdict (Figure~\ref{fig:tsnegrid}, bottom panel). Because t-SNE dimensions do not have intrinsic meanings, we treat these associations as descriptive patterns rather than definitive interpretations of these axes.

Cluster 3 ($n=98$) is the most interpretable. It is defined by near-universal Charges Refused (98\%), the highest Probable Cause Not Found rate (42\%), and near-zero rates for all plea and sentencing labels, including Pled Guilty to Charges and Trial Verdict Reached. This cluster captures cases terminated at the preliminary hearing stage, where the state’s charges are refused before the case reaches a plea, verdict, or sentencing stage. 

Cluster 0 ($n=75$) also shows a high Charges Refused rate (77\%) but retains some Probable Cause Found (28\%), suggesting cases where charges were ultimately refused after brief engagement with the court. Cluster 4 ($n=93$) has the highest Trial Verdict Reached rate (47\%), alongside high Pretrial Conference Held (84\%), Continuance Occurred (87\%), and Probable Cause Found (66\%). This cluster captures the small subset of cases that proceed through the full adversarial process to a verdict.

Cluster 5 ($n=85$) is distinguished by the highest Defendant In Custody And Not In Court rate (55\%), high Pled Guilty to Charges (78\%), and a high Sentence That Includes Jail Time rate (49\%). This profile reflects detained defendants who ultimately enter guilty pleas and receive sentences. Clusters 4 and 5 differ mostly in their disposition mode rather than how far a case progresses. Cluster 5 contains more cases in which the defendant pled guilty, whereas Cluster 4 contains more cases that reach a trial verdict. 

Clusters 1 ($n=81$) and 2 ($n=75$) share broadly similar profiles, including high Pled Not Guilty To Charges rates (98\% and 97\%), high Continuance Occurred rates (84\% and 77\%), and moderate-to-high sentencing rates. They differ in Pretrial Conference Held rates (90\% versus 40\%) and Pled Guilty to Charges rates (72\% versus 55\%). This suggests that Cluster 1 cases undergo more extensive pretrial proceedings before resolution.

\subsection{Outlier Analysis}
One value of this model to oversight is the ability to identify cases that deviate from model expectations. This can help surface potential procedural anomalies that warrant closer inspection. Here, we give examples of how this model could be used for monitoring the use of stay-away orders (SAOs). SAOs are operationally significant  because they have immediate behavioral consequences: they prohibit the defendant from contacting or approaching the protected party, and violation carries criminal penalties. Ensuring consistent application of SAOs thus has important safety implications.

In a ten proceeding case involving domestic abuse battery, domestic abuse second offense, and domestic abuse child endangerment, our model predicted a Domestic Stay-Away Order at Step 0 with very high confidence ($.995$), but there was no Stay Away Order recorded at this step. Given that these charges are squarely within the class of offenses for which a domestic SAO is routinely issued at first appearance in this jurisdiction, and that the second offense designation indicates a repeat-offender pattern already visible in the charge itself, a confident SAO prediction is a legally sensible prediction. The clerk note for Step 0 makes the source of the discrepancy explicit. The court wrote, ``The court could not issue a domestic stay away order in this matter as no victim information was provided to the court.'' The court was prepared to issue an SAO, however it was blocked due to no victim information available at the time of the hearing. Such cases can arise if, e.g., the police report does not contain victim information or if this information was not transmitted to the court in time for the hearing.

Similar patterns appear elsewhere: another high-confidence false positive ($.97$) involves a defendant charged with domestic abuse battery whose prior record includes additional domestic-violence offenses, where no SAO was issued at Step 0 because the defendant was never transported to court. This can occur when the defendant is already in jail but was not brought to court by the sheriff's office. Using the model to identify such discrepancies can help court observers better advocate for reforms that ensure procedural consistency.

\subsection{Integrated Gradients}

Since a major goal of the project is to support oversight rather than automate decision-making, interpretability is critical to our evaluation process.  To understand which input components drive model predictions, we use Integrated Gradients \cite{pmlr-v70-sundararajan17a} implemented through the Captum library~\cite{kokhlikyan2020captumunifiedgenericmodel}. 

The BERT text embedding achieves the highest mean raw attribution by a wide margin ($110.66 \pm 5.18$), followed by the case embedding ($41.97 \pm 1.86$) and  prior charge embedding ($18.77 \pm 2.44$). The BERT embedding attribution is highest for Pled Guilty to Charges ($137.03 \pm 2.13$) and Sentence That Does Not Include Jail ($133.63 \pm 6.48$). The case embedding similarly shows its strongest attributions for Pled Guilty to Charges ($59.42 \pm 1.71$) and Sentence That Does Not Include Jail Time Was Imposed ($56.82 \pm 2.46$), mirroring the BERT embedding's pattern for sentencing outcomes. (See Appendix~\ref{sec:ig} for more details.)

Integrated Gradients and ablation studies (Section~\ref{sec:ablation-studies}) independently identify clerk-note text as the dominant predictor, with structured case context (i.e., charges, prior labels, etc.) as a distant second, and defendant demographics, court identifiers, and bond/temporal scalars contributing to a lesser extent. Combined with the counterfactual results in Section~\ref{conterfactual-studies}, these consistent findings demonstrate that the model relies more on legally relevant content than demographic attributes.

\section{Conclusion}

Our results support three conclusions: (1)~proceeding-level predictions are feasible at scale, with the full transformer achieving a micro-F1 of $.5715$ and outperforming random and static MLP baselines and comparable to the GRU baseline; (2)~clerk note text is the dominant source of predictive signal, as removing proceeding text produces the largest single-component ablation drop and Integrated Gradients assigns the BERT text embedding the highest attribution scores; (3)~the model is more responsive to case-specific information than demographic attributes. Together, these findings show that proceeding-level modeling can move beyond static risk-style prediction toward a more dynamic account of how criminal cases unfold over time. They also suggest that interpretable sequence models can serve as useful tools for empirical court observation, helping researchers and court watchers identify patterns, anomalous cases, and points in the process where legal outcomes appear to shift. We emphasize that our empirical findings characterize the recorded practice of the single jurisdiction we study and should not be read as general claims about criminal courts. The proceeding-level modeling framework itself, however, is not jurisdiction-specific. Applying it elsewhere would require adapting the label extraction and features to local court data and retraining the model \citep{Wang2023}. 

\newpage
\section*{Limitations}
\label{sec:limitations}

One limitation of this analysis is that it relies upon text extraction from clerk notes to identify event outcome labels. While we have validated these through iterative review, errors in our extractors would influence the validity of our results. The reliance on regular expressions comes from the data infrastructure in most US court systems, which rarely gives clean, structured event information for public auditing. The state of a case is also determined by the data entry clerk, which introduces noise that caps the performance of any model trained on our data. Our manual validation confirms that the extracted labels are correct, but it cannot determine how many labels we miss. Each of the twenty labels appeared at least five times in the sample of 100 entries, and every extraction was correct. However, events whose phrasing our patterns fail to match are silently absent from the label set. These false negatives are unobserved by both the model and our evaluation and are treated in the training data as true negatives. Thus, this understates the prevalence of affected outcomes. 

Another limitation is generalizability beyond the jurisdiction studied. Although our modeling framework is jurisdiction-agnostic, the patterns identified in the paper reflect the practices of the criminal court system in a mid-sized U.S. city and should not be interpreted as general claims about criminal courts. Applying this elsewhere would require significant local adaptation. The regular expression extractors encode court-specific abbreviations, phrasing, and clerical conventions and would need to be rewritten to fit the local clerk minutes.

Additionally, the charge vocabulary, the Court A and Court B split by procedural stage, and section identifiers are specific to this jurisdiction and would need to be adapted for use elsewhere. The model would also need to be retrained on local data.

A related limitation extends beyond label extraction itself to the underlying records in which the label extractor operates. Clerk minutes and structured fields, such as the case identifier and charge codes, are entered by hand at the time of each proceeding, and our dataset inherits whatever transcription, attribution, and classification errors occur during this process. We have directly observed proceedings that appear in cases in which they do not belong, which is the result of a clerk entering a wrong case identifier when they enter their clerk notes, as well as proceeding text that references events or charges from unrelated matters. These kinds of errors are not unique to the jurisdiction studied in this paper. Because we have no independent ground truth against which to identify these errors at scale, the model is trained on a corpus that contains an unknown rate of cross-case contamination, miscoded charges, and incorrect procedure attributions, which sets a ceiling on achievable performance and may bias per-label metrics.

Another limitation is how the case embedding was set up, as we mapped charge descriptions to dense vectors. We apply a linear projection layer to a vocabulary count vector and initialize its weights randomly rather than using pre-trained semantic embeddings. As a result, the model treats semantically hierarchical charges, such as ``Possession of a Schedule 1 Drug'' or ``Possession of a Schedule 1 (2 Grams)'' as orthogonal and unrelated features. The model cannot infer that the specific charge with the dosage is a subclass of the general charge. This means that it must independently learn the consequences of a given charge from scratch. For rare, specific charges, such as the one above, the sparsity of the training data would mean that the model might fail to learn these weights effectively. 

A further limitation is that we do not report disaggregated evaluation \citep{10.1145/3461702.3462610}. Our counterfactual sensitivity analysis (Section~\ref{conterfactual-studies}) measures changes under demographic and case-attribute substitution, not whether error rates differ across defendants of different race, age, and gender. Thus, our finding that the model relies more on case characteristics rather than defendant demographics should be interpreted as a result about feature sensitivity and does not establish that performance is equitable across demographic groups. We leave such an evaluation to future work. 

Lastly, because we cannot redistribute the source records (Appendix~\ref{sec:dataandcodeavail}), our specific results are not independently reproducible without obtaining the underlying data from the court. The released pipeline and synthetic dataset allow inspection and rerunning of the method and label extraction, but not the numbers reported here.  
\newpage
\section*{Ethical Considerations}

\subsection*{Usage}

\subsubsection*{When To Use It}

The model is designed as an auditing and oversight tool for researchers, attorneys, reform advocates, and oversight bodies to better understand patterns and bias within the criminal court system's proceedings. It should be used on retrospective and aggregate analysis rather than individual case adjudication. For example, the model could be used to identify cases where outcomes deviated significantly from predictions, highlighting procedural irregularities or biases that occur for further investigation. It could also serve to measure policy improvement, seeing whether model predictions change after the establishment of a new policy and whether it achieves its intended effect. 

\subsubsection*{When Not to Use It}

This model should never be used as an automated decision-maker, as the system is designed to audit human decision-making rather than replacing it. It should not be used to make or influence decisions affecting bail, charging, plea, or sentencing for a defendant. If the model were to replace human judgment, the possible errors that would occur could directly harm defendants, while as an auditing tool, the errors could cause researchers to investigate the reasoning behind them.

\subsection*{Positive Impacts}

This research has the potential to increase transparency and accountability within the criminal court system of this jurisdiction by providing the tools that enable the researchers and oversight bodies to identify patterns in outcomes. Public defenders and legal aid organizations can utilize such tools to uncover systemic issues that affect their clients and, therefore, advocate for possible policy changes. Additionally, this research could empower journalists and researchers to conduct more research into the operations of this court system. 

\subsection*{Potential Harms}

Despite our intentions for this model to be used as an auditing tool, there could be some harms that come along with its deployment. The obvious risk is the potential of misuse, as discussed above. Another significant risk is privacy concerns as the model uses publicly available court records; the aggregation or analysis of data at scale could lead to the identification of individuals. 

\subsection*{Mitigation Strategies}

We adopt several mitigation strategies to limit the risks identified above. 

The first concerns restricting deployment to oversight rather than adjudication. Model outputs are framed throughout the paper as an auditing signal, not a recommendation. We will accompany any public release of the code with explicit documentation that prohibits use of the model to make, recommend, or score individual bail, charging, plea, or sentencing decisions.

The second strategy concerns data minimization. Although the underlying court records are publicly available, we do not redistribute them. Bulk release would allow for linkage of defendants, victims, and witnesses across cases and would defeat expungement relief that the defendants remain entitled to seek. We release code and a synthetic dataset instead. See Appendix~\ref{sec:dataandcodeavail}.

Third, we document the model's per-label performance and its weaker outcomes (Probable Cause Not Found, Charges Accepted, Bond Not Set), along with the data quality, extraction, generalizability, and evaluation limitations discussed in the Limitations section.

\section*{Acknowledgment}
A. Culotta was supported in part by NSF Awards IIS-HCC-2619509, DEB-2550002, CNS-2515250, IIS-HCC-2333537, CNS-2427237, the Louisiana Board of Regents Endowed Chairs for Eminent Scholars program, the Greenberg Family Professorship in Social Entrepreneurship, the Harold L. and Heather E. Jurist Center of Excellence for Artificial Intelligence at Tulane University, and the Tulane University Center of Excellence for Community-Engaged Artificial Intelligence.

\bibliography{custom, anthology-1, anthology-2}

\newpage

\appendix

\section{Extended Related Work}
\label{sec:related_work_full}

\subsection{Machine Learning for Single Decision Points in Criminal Justice}

The dominant research efforts in machine learning for criminal justice focus on predicting a single decision or outcome, such as pre-trial release, recidivism, or final case disposition. \citet{10.1093/qje/qjx032} train gradient-boosted decision trees on 758,027 New York City pretrial cases to forecast failure-to-appear and re-arrest, and found that algorithmic predictions could, in principle, reduce both jail populations and crime rates relative to judges. \citet{10.1145/3442188.3445908} takes an alternative approach at a different decision point, developing a hierarchical Bayesian model of cash bail assignments to disentangle judge-level variation from defendant-level risk while also addressing the infra-marginality problem inherent in observational bail data. 

The most widely known operational tool, COMPAS, was shown by \citet{angwin2016machinebias} to produce error rates that differed substantially across racial groups. In a closely related study, \citet{doi:10.1126/sciadv.aao5580} demonstrated that COMPAS is no more accurate than predictions made by laypersons with little or no criminal-justice expertise, and that a logistic regression using only two features matches its overall accuracy of roughly 65\%. \citet{Wang2023} shows that simple, interpretable models, in some cases compact enough to fit on a single page, can match black-box accuracy on Florida and Kentucky pretrial datasets while outperforming COMPAS, and they further document that recidivism models do not generalize well across jurisdictions and require local retraining. 

While this body of work has clarified the methodological foundations and limits of risk assessment, every study above collapses a case into a single prediction target, whether the defendant will reoffend, fail to appear, or receive cash bail, obscuring how decisions accumulate across the multiple proceedings that determine a defendant's trajectory through the system.

\subsection{Legal Judgment Prediction From Case Text}

Legal Judgment Prediction (LJP), a subtopic of NLP, uses textual descriptions of cases to forecast judicial outcomes. \citet{aletras2016predicting} introduced the task on the European Court of Human Rights and used bag-of-n-grams features and SVMs to predict whether a particular Convention article had been violated and achieved $\sim$ 79\% accuracy. \citet{chalkidis-etal-2019-neural} extended this benchmark to neural models on a larger European Court of Human Rights corpus and evaluated hierarchical attention networks and BERT-based architectures on binary, multi-label, and case-importance prediction. To accommodate the lengthy nature of legal documents, \citet{chalkidis-etal-2020-legal} pre-trained Legal-BERT on legislation, contracts, and court cases. \citet{niklaus-etal-2021-swiss} introduced the multilingual Swiss-Judgment-Prediction benchmark with hierarchical BERT variants. \citet{ma2021} explicitly moved beyond judge-summarized inputs by representing cases through their multi-stage life-cycle, but their target remains the final disposition. 

Recent work has expanded the LJP output structure beyond single-label classification. \citet{ZHANG2024112033} proposes HD-LJP, a hierarchical-dependency framework that jointly predicts law articles, charges, and terms of penalty on the CAIL2018 corpus by combining label-topological relation distillation with an ordered task-judicial-logic module in which each subtask's prediction filters the candidates for the next set. \citet{li-etal-2025-legal} formulates the prediction of law articles and charges as a multi-task multi-label problem and inserts label definitions and law-charge alignment knowledge into a transformer decoder model. While these methods relax the single-task or single-label assumption, the prediction target remains the terminal trial outcome of an already-completed case rather than the next event in its evolution.

Closest to our input structure is \citet{Polo_2021}, whose work classifies Brazilian legal proceedings into archived, active, or suspended status from chronological sequences of short clerk-written motions using LSTMs, MLP, and XGBoost classifiers over Word2Vec, Doc2Vec, TF-IDF, and BERT representations. Their setting is similar to ours in that they treat clerical text as a unit of observation and in modeling sequences sequentially. However, they collapse the proceedings into a single three-way status label, and we predict the multi-label event composition of each upcoming proceeding.  Across this literature, the prediction target is typically the terminal outcome of a case rather than the next event in its evolution. Our work shifts the unit of prediction to the individual proceeding.

\subsection{Sequence Models for Event Streams}

Most relevant to our methodological setup is work that adapts the Transformer architecture \cite{vaswani2017} to sequences of discrete categorical events accompanied by structured covariates. \citet{Li2020BEHRT} introduces a BERT-style transformer model, BEHRT, that represents a patient's electronic record as a sequence of diagnoses grouped into visits so that predictions are unaffected by the order of events within the same visit. Each diagnosis token's embedding sums disease code, age, position, and visit-segment embeddings. This model is pretrained on roughly 1.6 million UK primary-care records via masked language modeling and then fine-tuned to predict the diseases that will appear in a patient's next visit, next 6 months, and next 12 months. This model outperforms CNN and RNN baselines by 8-13\% in average precision. We view this model as a foundational model for health, as it is a single pre-trained representation that supports many downstream clinical prediction tasks. Another closely related model is CAREER \cite{vafa2024career}, a transformer-based foundational model that represents an individual's career as a sequence of jobs and predicts the next occupation, which is conditioned on prior employment history and demographics. CAREER is pre-trained on 24 million job sequences from resumes and is fine-tuned on smaller longitudinal surveys, namely the National Longitudinal Survey of Youth 1979 (NLSY79), its NLSY97 cohort, and the Panel Study of Income Dynamics (PSID), where it outperforms standard econometric baselines.

We treat a criminal case as a sequence of events with structure per-step covariates and learn to predict the next outcome. This applies the same broad recipe to the criminal-court setting. Our corpus is drawn from a single jurisdiction, so we stop short of claiming a foundational model for criminal court at the scale of BEHRT or CAREER. Our setup is otherwise analogous. However, our task differs from these prior efforts in mainly three ways. First, CAREER operates on a regular annual grid in which each timestep corresponds to a calendar year and BEHRT carries irregular timing through a patient-internal age embedding. But in our proceedings, they occur at irregular intervals, which we address through the use of discrete month-year positional encodings in addition to ordinal step indices. Secondly, our sequences are heterogeneously multi-modal, as we combine BERT-embedded clerk note text with structured demographic, charge, bond, and procedural stage features rather than just discrete occupation or diagnosis tokens with auxiliary covariates. Third, our supervision is multi-label per step, shared with BEHRT but
distinct from CAREER,  rather than single-label per step, since multiple events can occur within a single proceeding.

\section{Embedding Components}
\label{sec:embedding}
We describe each component of the proceeding representation $p_t^{(n)}$ below.

\paragraph{Text Embedding ($\psi_{\text{text}}$)}

Clerk notes text $a_t^{(n)}$ is encoded using a pre-trained BERT model \cite{devlin-etal-2019-bert}. The text is tokenized with the corresponding WordPiece tokenizer, truncated to 512 tokens, and then the CLS representation from the final layer is taken as the embedding. BERT is not fine-tuned and uses frozen weights. See Appendix~\ref{sec:implementation} for additional information on the BERT encoder.

\paragraph{Defendant Demographics ($\phi_{\text{demo}}$)}
Race (Black, White, Hispanic, Asian, Indigenous, Unknown), Gender (Male, Female, Unknown), and age at case initiation (continuous, in years) are combined into a single representation. Race and gender are encoded as learned embeddings, while age is normalized and concatenated directly. 

\paragraph{Case (Charge) Embedding ($\phi_{\text{charges}}$)}
Current charges for the case as of proceeding $t$ are represented as a vocabulary count vector $c_t^{(n)}$ over a fixed charge vocabulary $V$. A linear projection layer maps this sparse count vector to a dense embedding. The projection weights are therefore initialized randomly and learned from scratch during training. 

\paragraph{Prior Charge Embedding}
Aggregated prior charges $C_t^{(n)}$, which sum charge counts across all cases for the same defendant that began before proceeding $t$, are encoded with the
same $\phi_{\text{charges}}$. Current and prior charges therefore share projection weights and are two separate 192-dimensional blocks, distinguished only by their position in the concatenation.

\paragraph{Prior Label Embedding ($\phi_{\text{prior\_labels}}$)}
The multi-label outcome vector from the immediately preceding proceeding is passed through a linear projection. At $t = 1$, this vector is set to $\mathbf{0}^K$.

\paragraph{Court A Section ($\phi_{\text{court\_a\_sec}}$)}

The assigned Court A section identifier $d_t^{(n)}$ is encoded as a learned embedding and included only for proceedings that occur in Court A. The embedding is zeroed out for proceedings that take place in Court B.

\paragraph{Bond Amounts ($\psi_{\text{bond}}$)}
$j_t^{(n)}$ is the bond amount recorded at the defendant's initial hearing, while $k_t^{(n)}$ reflects the total amounts set or modified over the life of the case. Both are stored once per docket in the source court data and do not vary across proceedings within the same case. The two values are the same when the bond has never been revised and are different when it has been reset, revoked, or modified following new charges or violations. This allows the model to distinguish between a defendant whose bond has remained unchanged since the first appearance and one whose bond history reflects subsequent court action. Both values are encoded with the same deterministic encoder $\psi_{\text{bond}}$, which discretizes a dollar amount into one of the following bins: \{\$0, \$0--\$1,000, \$1,000--\$5,000, \$5,000--\$10,000, 
\$10,000--\$25,000, \$25,000--\$50,000, \$50,000--\$100,000,\\ 
$>\$100{,}000$, ``Not Set / Missing''\}. ``Not Set/ Missing'' is used both when the value is missing and at the sequence's initial state, where no bond has been observed yet. This returns a one-hot vector. 

\paragraph{Temporal Information ($\psi_{\text{time}}$)}

The month-year of each proceeding $\tau_t^{(n)}$ is encoded as a one-hot vector over all the month-year combinations observed in the training data. This discrete encoding complements the learned positional embedding $PE$, as $PE$ indexes ordinal position within the sequence, while the month-year one hot puts each step in terms of absolute calendar time and reflects the irregular spacing between proceedings. 

\paragraph{Scalar Features}

Two scalar indicators are concatenated directly to the feature vector without further transformation: the court-type indicator $h_t^{(n)}$ (0 for Court A, 1 for Court B) and the prior case count $l_t^{(n)}$, which counts the number of cases for the same defendant that precede this one.

\paragraph{Feature Fusion}
All components are concatenated along the feature dimension and passed through a single learned linear projection that maps the concatenated vector to the model's hidden dimension $d$, producing the final proceeding representation $p_t^{(n)}$

\section{Interpretability}
\label{sec:interp-appendix}

\paragraph{Integrated Gradients}
We report attributions for each of the eleven input component groups in Table~\ref{tab:appendix_raw_attribution_by_outcome}. Attributions are computed within each fold and averaged across folds. Because raw attribution magnitudes depend on component dimensionality, values for the 768-dimensional BERT embedding are not directly comparable to those for scalar features such as prior case count. We therefore also report z-scores computed within each outcome (Table~\ref{tab:appendix_within_label_zscore_by_outcome}), showing which inputs drive a given prediction, and across outcomes (Table~\ref{tab:appendix_across_label_zscore_by_outcome}), showing which outcomes a given input is most influential for. Outcomes present in only a single fold have an undefined standard deviation and are marked $\mathrm{NA}$. 

\begin{figure}[t]
    \centering
    \includegraphics[width=1\linewidth]{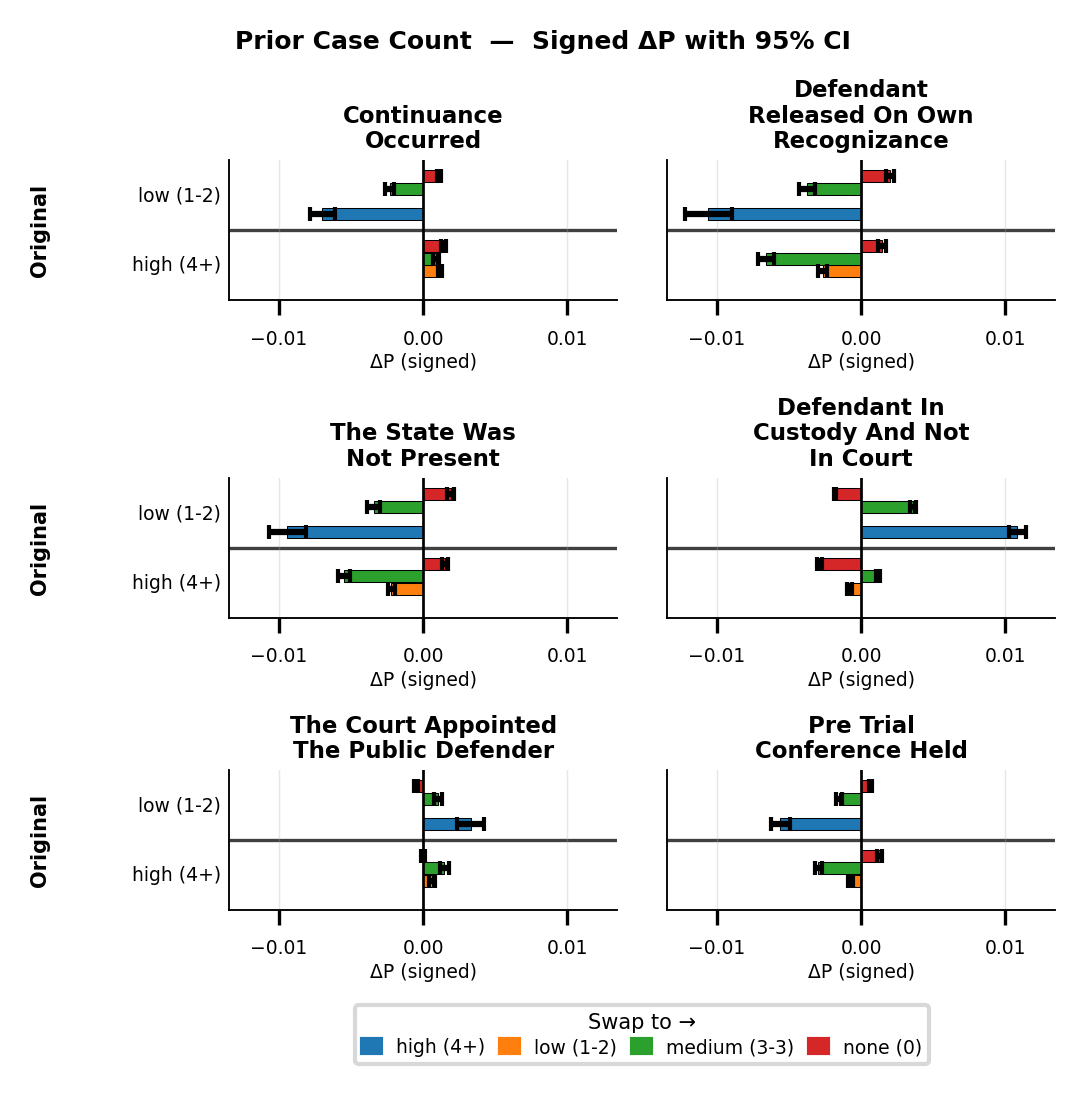}
    \caption{Prior Case Count Counterfactual}
    \label{fig:pccconterfactual}
\end{figure}

\begin{figure}[t]
    \centering
    \includegraphics[width=1\linewidth]{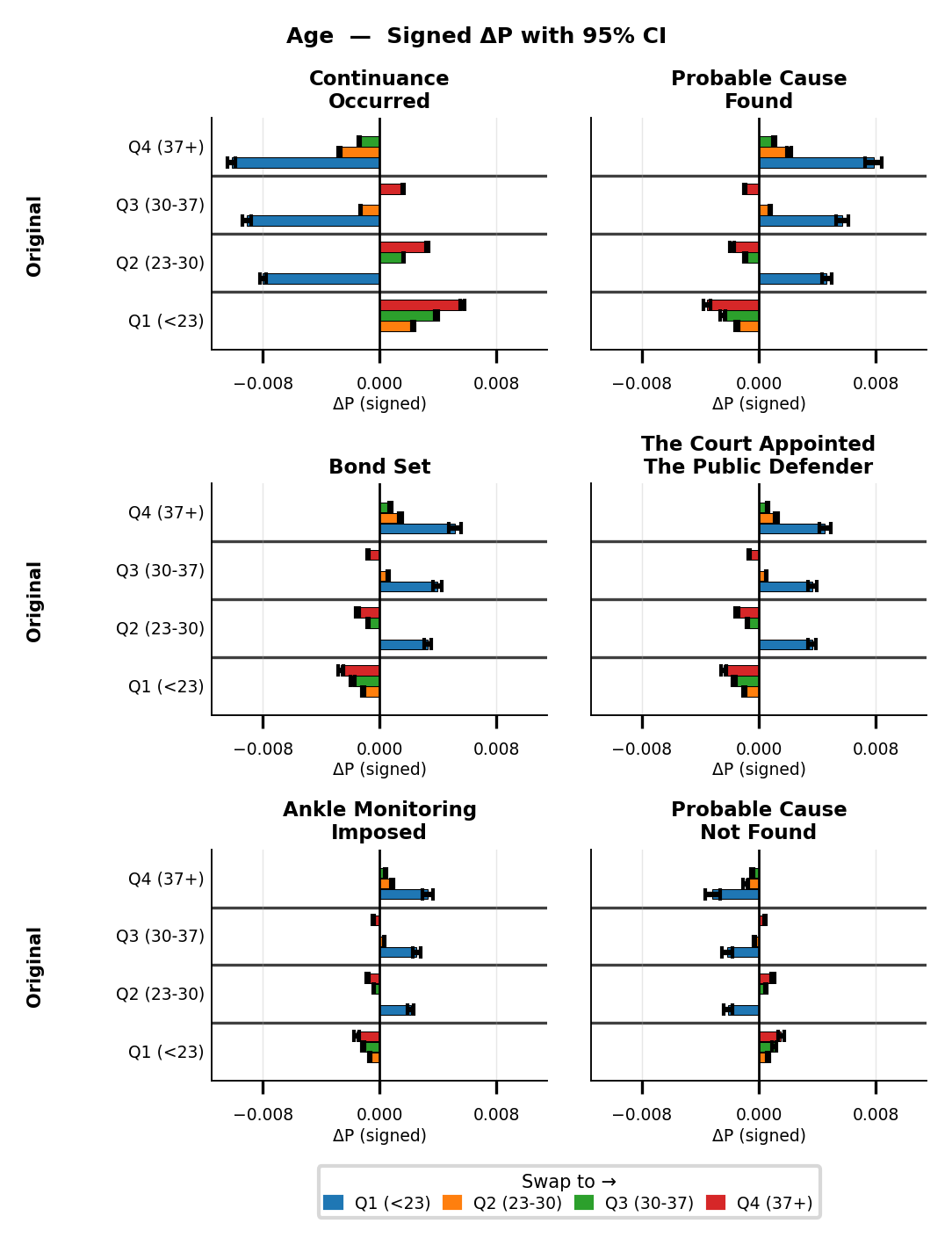}
    \caption{Age Counterfactual}
    \label{fig:agecounterfactual}
\end{figure}

\begin{figure}[t]
    \centering
    \includegraphics[width=1\linewidth]{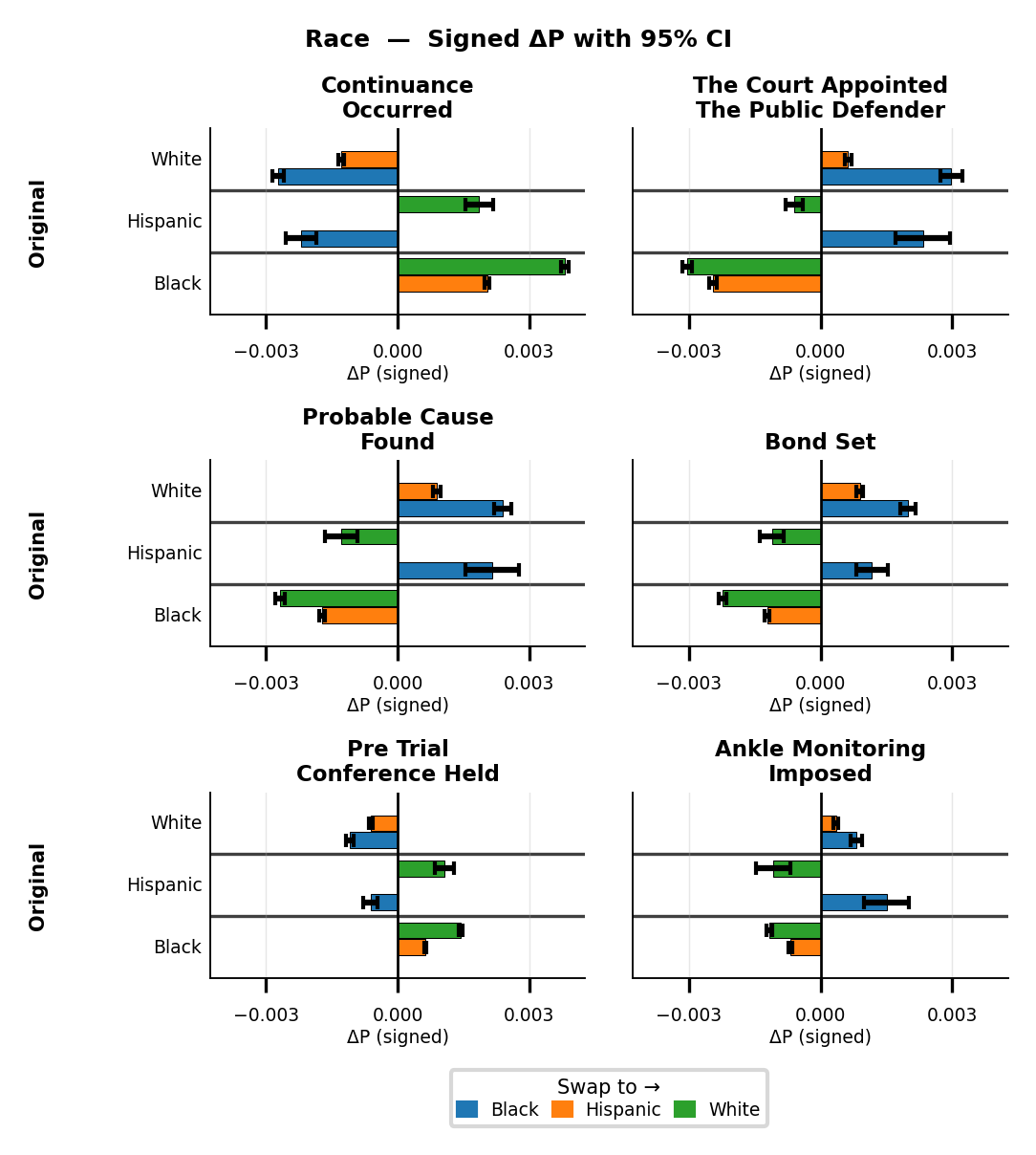}
    \caption{Race Counterfactual}
    \label{fig:racecounterfactual}
\end{figure}

\paragraph{Counterfactual Sensitivity}
Counterfactual fairness in its strict form requires a fully specified causal graph. Our procedure, which is the process of swapping each demographic value to all other observed values and measuring the resulting shift in predicted probability, provides an empirical proxy that does not require causal identification, while still allowing us to compare the model's responsiveness to legally relevant features (e.g., prior case count) against protected ones (e.g., race). Full per-label breakdowns for prior case count, age, and race appear in Figures~\ref{fig:pccconterfactual}--\ref{fig:racecounterfactual}.

\section{Proceeding-Level Per-Label Performance}

Table~\ref{tab:per_label_transformer_2_appendix} reports the proceeding-level per-label performance for the full transformer under 5-fold cross validation. Performance is evaluated at each individual proceeding against the ground-truth labels (regex-generated) for that step.

\begin{table*}[htbp]
\centering
\small
\caption{Proceeding-level per-label performance (mean $\pm$ std over 5 folds), sorted by F1 score.}
\label{tab:per_label_transformer_2_appendix}
\begin{tabular}{lcccc}
\toprule
Label & Precision & Recall & F1-Score & Support \\
\midrule
First Appearance Hearing Occurred & $.9977 \pm .00$ & $.9877 \pm .01$ & $.9926 \pm .00$ & $517.8 \pm 3.5$ \\
Court Appointed Public Defender & $.7992 \pm .02$ & $.7961 \pm .02$ & $.7972 \pm .01$ & $634.6 \pm 11.3$ \\
Bond Set & $.7123 \pm .02$ & $.7598 \pm .01$ & $.7351 \pm .01$ & $740.6 \pm 14.9$ \\
Pre-Trial Conference Held & $.6598 \pm .01$ & $.6563 \pm .03$ & $.6576 \pm .02$ & $625.6 \pm 73.0$ \\
Domestic Stay-Away Order Imposed & $.6350 \pm .03$ & $.6657 \pm .05$ & $.6479 \pm .02$ & $109.2 \pm 4.0$ \\
Pled Not Guilty to Charges & $.5489 \pm .03$ & $.7883 \pm .02$ & $.6469 \pm .03$ & $326.6 \pm 7.9$ \\
Charges Refused & $.4941 \pm .03$ & $.4958 \pm .02$ & $.4939 \pm .01$ & $369.6 \pm 13.3$ \\
State Was Not Present & $.6098 \pm .07$ & $.4118 \pm .05$ & $.4914 \pm .06$ & $137.0 \pm 13.7$ \\
Probable Cause Found & $.5741 \pm .03$ & $.4054 \pm .03$ & $.4743 \pm .02$ & $286.2 \pm 8.6$ \\
Defendant Released on Own Recognizance & $.5242 \pm .04$ & $.4060 \pm .07$ & $.4549 \pm .05$ & $179.8 \pm 18.9$ \\
Continuance Occurred & $.4452 \pm .01$ & $.4415 \pm .04$ & $.4426 \pm .02$ & $1084.8 \pm 90.1$ \\
Ankle Monitoring Imposed & $.5673 \pm .07$ & $.3571 \pm .06$ & $.4376 \pm .06$ & $91.2 \pm 6.5$ \\
Sentence Without Jail Time Imposed & $.4596 \pm .06$ & $.3272 \pm .04$ & $.3793 \pm .04$ & $80.2 \pm 8.4$ \\
Pled Guilty to Charges & $.2939 \pm .04$ & $.5214 \pm .03$ & $.3733 \pm .03$ & $215.4 \pm 5.7$ \\
Sentence Including Jail Time Imposed & $.2851 \pm .05$ & $.4207 \pm .04$ & $.3388 \pm .04$ & $161.2 \pm 7.8$ \\
Defendant in Custody and Not in Court & $.4016 \pm .02$ & $.2496 \pm .03$ & $.3075 \pm .03$ & $338.6 \pm 46.9$ \\
Trial Verdict Reached & $.2800 \pm .03$ & $.2975 \pm .07$ & $.2838 \pm .04$ & $80.8 \pm 9.0$ \\
Probable Cause Not Found & $.2349 \pm .03$ & $.2362 \pm .05$ & $.2349 \pm .04$ & $88.4 \pm 3.6$ \\
Bond Not Set & $.2206 \pm .04$ & $.1349 \pm .01$ & $.1661 \pm .01$ & $198.2 \pm 22.0$ \\
Charges Accepted & $.1425 \pm .01$ & $.1305 \pm .04$ & $.1340 \pm .02$ & $92.4 \pm 9.1$ \\
\bottomrule
\end{tabular}
\end{table*}

\section{Case-Level Evaluation}

Table~\ref{tab:per_class_case_level_sorted} reports case-level per-label performance, where proceeding-level predictions are aggregated across all proceedings within a case. Performance is consistently higher than at the proceeding level, as the model has numerous chances across a case to correctly identify whether each outcome ever occurs. Outcomes such as ``Continuance Occurred'' and ``Pled Guilty to Charges'' show the largest gains from aggregation, which suggests that these outcomes have distributed lexical indicators that accumulate across proceedings.

\begin{table*}[htbp]
\centering
\caption{Per-Class Performance for Case Level Evaluation (Averaged over 5 Folds), Sorted by F1-Score}
\label{tab:per_class_case_level_sorted}
\small
\begin{tabular}{lcccc}
\toprule
Label & Precision & Recall & F1-Score & Support \\
\midrule
First Appearance Hearing Occurred & $.999 \pm .001$ & $.999 \pm .001$ & $.999 \pm .001$ & $506.2 \pm 0.837$ \\
Bond Set & $.950 \pm .013$ & $.948 \pm .009$ & $.949 \pm .004$ & $463.6 \pm 7.701$ \\
Court Appointed Public Defender & $.894 \pm .017$ & $.924 \pm .026$ & $.909 \pm .013$ & $415.6 \pm 6.841$ \\
Pled Not Guilty to Charges & $.902 \pm .020$ & $.829 \pm .025$ & $.864 \pm .010$ & $318.2 \pm 7.791$ \\
Pre-Trial Conference Held & $.824 \pm .028$ & $.842 \pm .016$ & $.833 \pm .016$ & $227.0 \pm 16.778$ \\
Continuance Occurred & $.778 \pm .012$ & $.755 \pm .030$ & $.766 \pm .017$ & $329.2 \pm 2.588$ \\
Domestic Stay-Away Order Imposed & $.650 \pm .032$ & $.689 \pm .055$ & $.667 \pm .016$ & $107.0 \pm 4.472$ \\
Charges Refused & $.829 \pm .027$ & $.551 \pm .030$ & $.662 \pm .017$ & $293.4 \pm 14.311$ \\
Pled Guilty to Charges & $.798 \pm .033$ & $.538 \pm .033$ & $.642 \pm .026$ & $212.6 \pm 7.603$ \\
Probable Cause Found & $.733 \pm .021$ & $.550 \pm .039$ & $.628 \pm .027$ & $232.0 \pm 6.782$ \\
Defendant Released on Own Recognizance & $.646 \pm .035$ & $.542 \pm .051$ & $.588 \pm .030$ & $112.4 \pm 12.095$ \\
State Was Not Present & $.665 \pm .070$ & $.495 \pm .080$ & $.567 \pm .077$ & $106.4 \pm 10.784$ \\
Defendant in Custody and Not in Court & $.680 \pm .016$ & $.479 \pm .043$ & $.561 \pm .027$ & $121.6 \pm 7.301$ \\
Sentence Including Jail Time Imposed & $.668 \pm .016$ & $.438 \pm .052$ & $.527 \pm .042$ & $161.2 \pm 8.758$ \\
Ankle Monitoring Imposed & $.673 \pm .077$ & $.421 \pm .059$ & $.518 \pm .065$ & $83.4 \pm 7.436$ \\
Sentence Without Jail Time Imposed & $.643 \pm .070$ & $.342 \pm .061$ & $.444 \pm .058$ & $80.2 \pm 9.365$ \\
Trial Verdict Reached & $.602 \pm .047$ & $.317 \pm .064$ & $.411 \pm .055$ & $80.0 \pm 10.124$ \\
Bond Not Set & $.462 \pm .043$ & $.267 \pm .023$ & $.337 \pm .023$ & $156.2 \pm 16.407$ \\
Probable Cause Not Found & $.284 \pm .026$ & $.279 \pm .036$ & $.281 \pm .030$ & $84.2 \pm 4.970$ \\
Charges Accepted & $.330 \pm .049$ & $.226 \pm .056$ & $.264 \pm .045$ & $90.8 \pm 10.402$ \\
\bottomrule
\end{tabular}
\end{table*}

\section{K-means}

\begin{figure}[t]
    \includegraphics[width=1\linewidth]{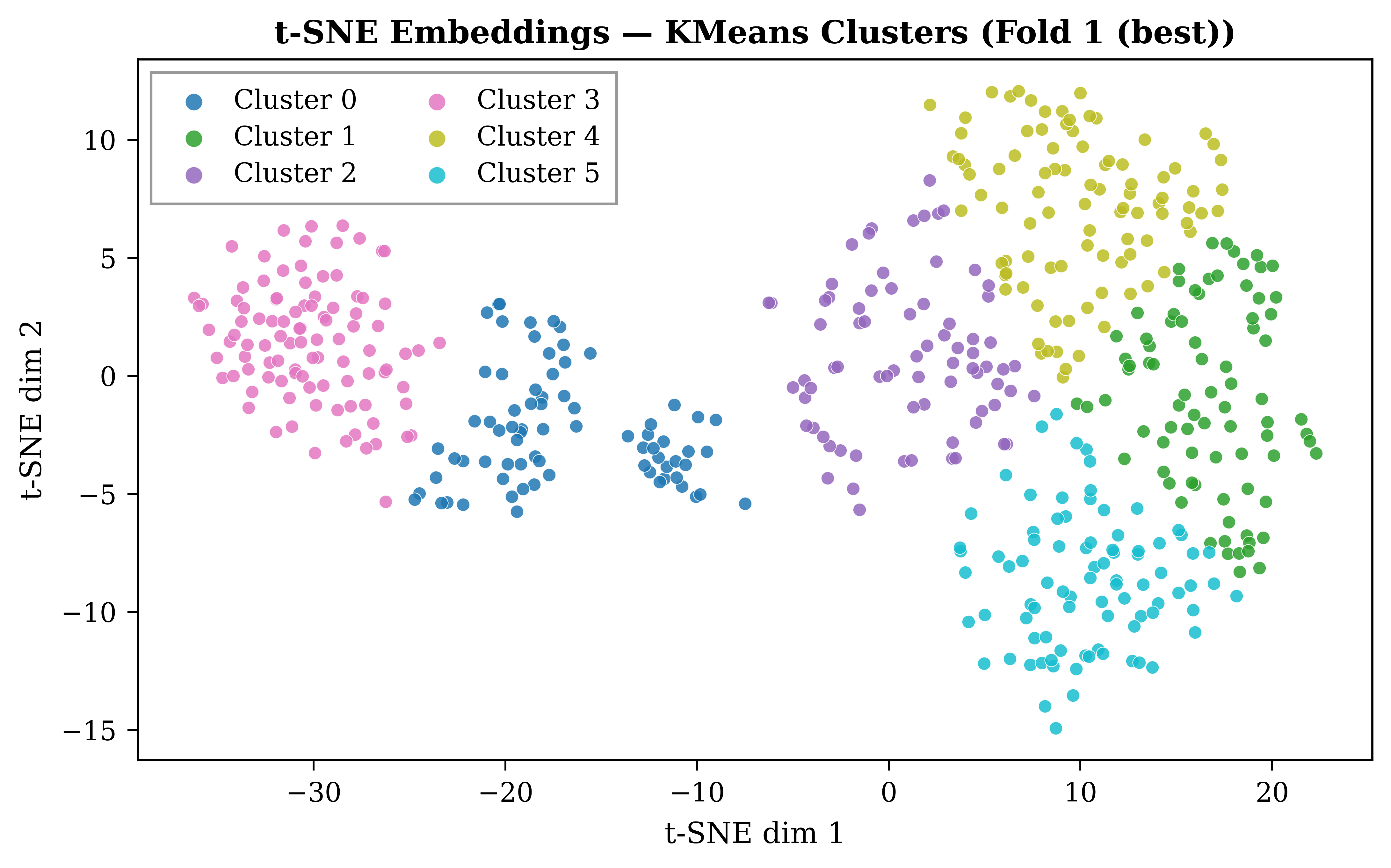}
    \caption{Clustering results}
    \label{fig:tsneclusters}
\end{figure}

Figure~\ref{fig:tsneclusters} shows the results of t-SNE clustering.

\section{Attribution Tables}
\label{sec:ig}

Tables~\ref{tab:appendix_raw_attribution_by_outcome}--\ref{tab:appendix_across_label_zscore_by_outcome} report the fold-level attribution summaries used in the feature attribution analysis. Each cell reports the mean and standard deviation across available folds. Cells with $\mathrm{NA}$ have an undefined standard deviation because the outcome was present in only one fold.

\begin{table*}[t]
\centering
\setlength{\tabcolsep}{3pt}
\renewcommand{\arraystretch}{1.15}
\caption{Raw attribution magnitude by outcome and feature group. Cells report mean $\pm$ standard deviation across folds. Outcome names are abbreviated. See Table~\ref{tab:label_distribution} for full label names.}
\label{tab:appendix_raw_attribution_by_outcome}
\resizebox{\textwidth}{!}{%
\begin{tabular}{@{}l*{11}{c}@{}}
\toprule
Outcome & Text emb. & Case emb. & Prior charge emb. & Prior label emb. & Court A Sec. emb. & Def. emb. & Temporal & Bond sum & Orig. bond & Court type & Prior cases \\
\midrule
Bond Set & $110.8050 \pm \mathrm{NA}$ & $30.3496 \pm \mathrm{NA}$ & $23.3237 \pm \mathrm{NA}$ & $10.9694 \pm \mathrm{NA}$ & $4.4393 \pm \mathrm{NA}$ & $2.6275 \pm \mathrm{NA}$ & $0.7764 \pm \mathrm{NA}$ & $1.8679 \pm \mathrm{NA}$ & $0.7430 \pm \mathrm{NA}$ & $1.2260 \pm \mathrm{NA}$ & $0.4723 \pm \mathrm{NA}$ \\
Public Defender Appointed & $113.5933 \pm 15.552$ & $47.4977 \pm 7.576$ & $15.2448 \pm 11.162$ & $13.7651 \pm 2.373$ & $4.7651 \pm 0.676$ & $4.2359 \pm 0.761$ & $1.7981 \pm 0.201$ & $1.3785 \pm 0.540$ & $0.6102 \pm 0.221$ & $0.4776 \pm 0.297$ & $0.1670 \pm 0.166$ \\
Released on Recognizance & $68.9643 \pm 12.285$ & $42.2989 \pm 7.870$ & $14.5345 \pm 14.089$ & $5.3485 \pm 0.807$ & $3.0551 \pm 1.402$ & $3.1483 \pm 0.472$ & $1.0612 \pm 0.252$ & $1.0387 \pm 0.519$ & $0.5263 \pm 0.248$ & $0.2027 \pm 0.129$ & $0.2432 \pm 0.291$ \\
Ankle Monitoring Imposed & $77.5265 \pm 13.980$ & $28.8517 \pm 4.428$ & $22.9491 \pm 12.048$ & $8.1434 \pm 1.590$ & $3.3123 \pm 0.711$ & $2.6922 \pm 1.132$ & $0.8011 \pm 0.200$ & $0.7985 \pm 0.539$ & $0.5078 \pm 0.287$ & $0.5243 \pm 0.358$ & $0.2644 \pm 0.278$ \\
Charges Refused & $78.9577 \pm 9.130$ & $30.3561 \pm 3.820$ & $11.3440 \pm 2.651$ & $7.0821 \pm 0.654$ & $3.0278 \pm 0.643$ & $2.7769 \pm 0.261$ & $0.8894 \pm 0.104$ & $2.0752 \pm 0.512$ & $0.3233 \pm 0.041$ & $0.3347 \pm 0.074$ & $0.2609 \pm 0.072$ \\
Probable Cause Found & $111.6777 \pm \mathrm{NA}$ & $34.3002 \pm \mathrm{NA}$ & $29.7379 \pm \mathrm{NA}$ & $9.1841 \pm \mathrm{NA}$ & $2.5287 \pm \mathrm{NA}$ & $3.0105 \pm \mathrm{NA}$ & $1.4470 \pm \mathrm{NA}$ & $0.3689 \pm \mathrm{NA}$ & $0.3179 \pm \mathrm{NA}$ & $0.2181 \pm \mathrm{NA}$ & $0.2004 \pm \mathrm{NA}$ \\
Pretrial Conference Held & $135.9125 \pm 7.196$ & $36.1745 \pm 2.054$ & $17.5502 \pm 1.944$ & $10.2318 \pm 0.726$ & $3.8323 \pm 0.695$ & $3.2582 \pm 0.243$ & $0.8663 \pm 0.098$ & $0.9027 \pm 0.067$ & $0.5989 \pm 0.228$ & $0.4041 \pm 0.051$ & $0.4952 \pm 0.169$ \\
Pled Not Guilty & $87.1481 \pm 23.588$ & $25.7732 \pm 4.463$ & $24.2643 \pm 22.452$ & $10.4375 \pm 2.641$ & $3.2995 \pm 0.407$ & $2.4129 \pm 0.091$ & $0.9035 \pm 0.277$ & $0.6346 \pm 0.101$ & $0.5222 \pm 0.246$ & $0.5374 \pm 0.168$ & $0.4823 \pm 0.350$ \\
Continuance Occurred & $99.6486 \pm 13.705$ & $33.7928 \pm 3.335$ & $14.8666 \pm 7.772$ & $8.4613 \pm 1.340$ & $2.8407 \pm 0.420$ & $2.7832 \pm 0.387$ & $0.6229 \pm 0.070$ & $0.9160 \pm 0.228$ & $0.5323 \pm 0.150$ & $0.3534 \pm 0.147$ & $0.2535 \pm 0.128$ \\
In Custody, Not in Court & $97.2555 \pm 13.158$ & $34.7989 \pm 9.306$ & $18.4545 \pm 26.097$ & $7.8922 \pm 0.425$ & $2.9123 \pm 0.125$ & $3.5423 \pm 0.590$ & $0.6070 \pm 0.119$ & $0.9809 \pm 0.393$ & $0.3108 \pm 0.217$ & $0.3091 \pm 0.175$ & $0.1495 \pm 0.211$ \\
Pled Guilty & $137.0264 \pm 2.133$ & $59.4222 \pm 1.714$ & $24.1409 \pm 1.334$ & $11.4992 \pm 0.587$ & $5.7652 \pm 1.770$ & $4.8484 \pm 0.502$ & $1.3225 \pm 0.202$ & $1.3847 \pm 0.088$ & $0.8151 \pm 0.219$ & $0.9314 \pm 0.309$ & $0.7666 \pm 0.065$ \\
Trial Verdict Reached & $115.8153 \pm 7.414$ & $39.2843 \pm 2.659$ & $17.5370 \pm 4.967$ & $8.4057 \pm 0.487$ & $3.5867 \pm 0.666$ & $3.5114 \pm 0.493$ & $0.9982 \pm 0.144$ & $0.8832 \pm 0.171$ & $0.5183 \pm 0.114$ & $0.6924 \pm 0.272$ & $0.4373 \pm 0.182$ \\
Sentence Includes Jail Time & $130.3489 \pm 5.197$ & $51.9305 \pm 2.888$ & $26.6803 \pm 1.786$ & $10.4391 \pm 0.831$ & $4.8257 \pm 1.476$ & $4.6398 \pm 0.802$ & $1.0136 \pm 0.071$ & $1.3005 \pm 0.277$ & $0.7442 \pm 0.240$ & $0.8327 \pm 0.255$ & $0.6768 \pm 0.099$ \\
Sentence Excludes Jail Time & $133.6322 \pm 6.479$ & $56.8237 \pm 2.458$ & $18.4084 \pm 3.461$ & $10.0458 \pm 0.330$ & $5.5110 \pm 1.109$ & $4.3974 \pm 0.230$ & $1.8706 \pm 0.278$ & $1.2729 \pm 0.194$ & $0.7076 \pm 0.092$ & $0.7578 \pm 0.241$ & $0.7206 \pm 0.143$ \\
Mean Attribution & $110.6558 \pm 5.184$ & $41.9691 \pm 1.855$ & $18.7662 \pm 2.438$ & $9.5488 \pm 0.496$ & $4.0503 \pm 0.720$ & $3.5957 \pm 0.309$ & $1.1170 \pm 0.113$ & $1.1852 \pm 0.144$ & $0.5860 \pm 0.136$ & $0.5630 \pm 0.042$ & $0.4286 \pm 0.045$ \\
\bottomrule
\end{tabular}%
}
\end{table*}

\begin{table*}[t]
\centering
\setlength{\tabcolsep}{3pt}
\renewcommand{\arraystretch}{1.15}
\caption{Within-label standardized attribution by outcome and feature group. Cells report mean $\pm$ standard deviation across folds.}
\label{tab:appendix_within_label_zscore_by_outcome}
\resizebox{\textwidth}{!}{%
\begin{tabular}{@{}l*{11}{c}@{}}
\toprule
Outcome & Text emb. & Case emb. & Prior charge emb. & Prior label emb. & Court A Sec. emb. & Def. emb. & Temporal & Bond sum & Orig. bond & Court type & Prior cases \\
\midrule
Bond Set & $3.0064 \pm \mathrm{NA}$ & $0.4263 \pm \mathrm{NA}$ & $0.2010 \pm \mathrm{NA}$ & $-0.1951 \pm \mathrm{NA}$ & $-0.4045 \pm \mathrm{NA}$ & $-0.4626 \pm \mathrm{NA}$ & $-0.5220 \pm \mathrm{NA}$ & $-0.4870 \pm \mathrm{NA}$ & $-0.5231 \pm \mathrm{NA}$ & $-0.5076 \pm \mathrm{NA}$ & $-0.5318 \pm \mathrm{NA}$ \\
Public Defender Appointed & $2.8787 \pm 0.035$ & $0.8750 \pm 0.117$ & $-0.0763 \pm 0.355$ & $-0.1465 \pm 0.047$ & $-0.4178 \pm 0.019$ & $-0.4339 \pm 0.042$ & $-0.5072 \pm 0.026$ & $-0.5217 \pm 0.033$ & $-0.5444 \pm 0.031$ & $-0.5487 \pm 0.035$ & $-0.5572 \pm 0.024$ \\
Released on Recognizance & $2.5751 \pm 0.249$ & $1.3608 \pm 0.182$ & $0.1165 \pm 0.677$ & $-0.3443 \pm 0.077$ & $-0.4527 \pm 0.071$ & $-0.4454 \pm 0.072$ & $-0.5407 \pm 0.064$ & $-0.5414 \pm 0.097$ & $-0.5665 \pm 0.073$ & $-0.5815 \pm 0.079$ & $-0.5800 \pm 0.074$ \\
Ankle Monitoring Imposed & $2.8643 \pm 0.055$ & $0.7016 \pm 0.087$ & $0.3845 \pm 0.365$ & $-0.2264 \pm 0.030$ & $-0.4349 \pm 0.082$ & $-0.4693 \pm 0.050$ & $-0.5532 \pm 0.028$ & $-0.5551 \pm 0.014$ & $-0.5669 \pm 0.021$ & $-0.5669 \pm 0.015$ & $-0.5778 \pm 0.024$ \\
Charges Refused & $2.9323 \pm 0.032$ & $0.7871 \pm 0.062$ & $-0.0438 \pm 0.119$ & $-0.2387 \pm 0.018$ & $-0.4176 \pm 0.039$ & $-0.4291 \pm 0.019$ & $-0.5128 \pm 0.017$ & $-0.4615 \pm 0.026$ & $-0.5379 \pm 0.014$ & $-0.5375 \pm 0.015$ & $-0.5405 \pm 0.013$ \\
Probable Cause Found & $2.9436 \pm \mathrm{NA}$ & $0.5240 \pm \mathrm{NA}$ & $0.3813 \pm \mathrm{NA}$ & $-0.2614 \pm \mathrm{NA}$ & $-0.4696 \pm \mathrm{NA}$ & $-0.4545 \pm \mathrm{NA}$ & $-0.5034 \pm \mathrm{NA}$ & $-0.5371 \pm \mathrm{NA}$ & $-0.5387 \pm \mathrm{NA}$ & $-0.5418 \pm \mathrm{NA}$ & $-0.5424 \pm \mathrm{NA}$ \\
Pretrial Conference Held & $3.0427 \pm 0.005$ & $0.4448 \pm 0.037$ & $-0.0406 \pm 0.045$ & $-0.2316 \pm 0.009$ & $-0.3975 \pm 0.019$ & $-0.4128 \pm 0.006$ & $-0.4754 \pm 0.005$ & $-0.4745 \pm 0.005$ & $-0.4825 \pm 0.008$ & $-0.4875 \pm 0.004$ & $-0.4851 \pm 0.003$ \\
Pled Not Guilty & $2.9299 \pm 0.094$ & $0.4848 \pm 0.159$ & $0.3147 \pm 0.609$ & $-0.1438 \pm 0.053$ & $-0.4281 \pm 0.061$ & $-0.4631 \pm 0.063$ & $-0.5272 \pm 0.037$ & $-0.5374 \pm 0.042$ & $-0.5429 \pm 0.034$ & $-0.5418 \pm 0.038$ & $-0.5452 \pm 0.029$ \\
Continuance Occurred & $2.9649 \pm 0.028$ & $0.6621 \pm 0.057$ & $0.0024 \pm 0.251$ & $-0.2312 \pm 0.021$ & $-0.4268 \pm 0.027$ & $-0.4291 \pm 0.023$ & $-0.5055 \pm 0.026$ & $-0.4954 \pm 0.029$ & $-0.5085 \pm 0.028$ & $-0.5151 \pm 0.030$ & $-0.5179 \pm 0.023$ \\
In Custody, Not in Court & $2.8707 \pm 0.097$ & $0.6710 \pm 0.248$ & $0.1745 \pm 0.929$ & $-0.2581 \pm 0.028$ & $-0.4355 \pm 0.070$ & $-0.4142 \pm 0.084$ & $-0.5164 \pm 0.072$ & $-0.5044 \pm 0.089$ & $-0.5277 \pm 0.086$ & $-0.5276 \pm 0.084$ & $-0.5324 \pm 0.071$ \\
Pled Guilty & $2.8702 \pm 0.009$ & $0.9244 \pm 0.027$ & $0.0406 \pm 0.040$ & $-0.2767 \pm 0.013$ & $-0.4208 \pm 0.040$ & $-0.4434 \pm 0.009$ & $-0.5319 \pm 0.006$ & $-0.5303 \pm 0.005$ & $-0.5447 \pm 0.009$ & $-0.5417 \pm 0.007$ & $-0.5458 \pm 0.005$ \\
Trial Verdict Reached & $2.9734 \pm 0.013$ & $0.6609 \pm 0.052$ & $0.0037 \pm 0.135$ & $-0.2721 \pm 0.016$ & $-0.4180 \pm 0.020$ & $-0.4201 \pm 0.012$ & $-0.4966 \pm 0.015$ & $-0.5002 \pm 0.015$ & $-0.5112 \pm 0.019$ & $-0.5061 \pm 0.017$ & $-0.5136 \pm 0.015$ \\
Sentence Includes Jail Time & $2.8968 \pm 0.010$ & $0.8145 \pm 0.031$ & $0.1461 \pm 0.055$ & $-0.2865 \pm 0.016$ & $-0.4345 \pm 0.038$ & $-0.4401 \pm 0.015$ & $-0.5366 \pm 0.009$ & $-0.5292 \pm 0.013$ & $-0.5438 \pm 0.013$ & $-0.5412 \pm 0.007$ & $-0.5455 \pm 0.011$ \\
Sentence Excludes Jail Time & $2.8905 \pm 0.020$ & $0.9152 \pm 0.053$ & $-0.0737 \pm 0.083$ & $-0.2890 \pm 0.015$ & $-0.4061 \pm 0.031$ & $-0.4346 \pm 0.007$ & $-0.4996 \pm 0.008$ & $-0.5152 \pm 0.009$ & $-0.5298 \pm 0.007$ & $-0.5284 \pm 0.012$ & $-0.5294 \pm 0.007$ \\
Mean Z-Score Within Label & $2.9046 \pm 0.028$ & $0.7714 \pm 0.069$ & $0.0535 \pm 0.063$ & $-0.2474 \pm 0.013$ & $-0.4218 \pm 0.025$ & $-0.4357 \pm 0.014$ & $-0.5140 \pm 0.005$ & $-0.5106 \pm 0.009$ & $-0.5312 \pm 0.006$ & $-0.5323 \pm 0.008$ & $-0.5365 \pm 0.007$ \\
\bottomrule
\end{tabular}%
}
\end{table*}

\begin{table*}[t]
\centering
\setlength{\tabcolsep}{3pt}
\renewcommand{\arraystretch}{1.15}
\caption{Across-label standardized attribution by outcome and feature group. Cells report mean $\pm$ standard deviation across folds.}
\label{tab:appendix_across_label_zscore_by_outcome}
\resizebox{\textwidth}{!}{%
\begin{tabular}{@{}l*{11}{c}@{}}
\toprule
Outcome & Text emb. & Case emb. & Prior charge emb. & Prior label emb. & Court A Sec. emb. & Def. emb. & Temporal & Bond sum & Orig. bond & Court type & Prior cases \\
\midrule
Bond Set & $-0.3007 \pm \mathrm{NA}$ & $-1.1217 \pm \mathrm{NA}$ & $0.2522 \pm \mathrm{NA}$ & $0.8114 \pm \mathrm{NA}$ & $0.4422 \pm \mathrm{NA}$ & $-0.6134 \pm \mathrm{NA}$ & $-0.7730 \pm \mathrm{NA}$ & $1.4749 \pm \mathrm{NA}$ & $-0.2053 \pm \mathrm{NA}$ & $2.3795 \pm \mathrm{NA}$ & $-0.0066 \pm \mathrm{NA}$ \\
Public Defender Appointed & $0.0955 \pm 0.598$ & $0.5233 \pm 0.666$ & $-0.3610 \pm 1.152$ & $1.8460 \pm 0.599$ & $0.8723 \pm 1.006$ & $0.7247 \pm 0.567$ & $1.6455 \pm 0.343$ & $0.3419 \pm 0.968$ & $0.1299 \pm 1.266$ & $-0.2940 \pm 1.000$ & $-0.9960 \pm 0.524$ \\
Released on Recognizance & $-1.5175 \pm 0.616$ & $-0.0076 \pm 0.747$ & $-0.2740 \pm 1.653$ & $-1.5742 \pm 0.277$ & $-0.9152 \pm 0.528$ & $-0.6113 \pm 0.524$ & $-0.2969 \pm 0.516$ & $-0.1974 \pm 1.064$ & $-0.2548 \pm 1.099$ & $-1.0967 \pm 0.555$ & $-0.6302 \pm 1.012$ \\
Ankle Monitoring Imposed & $-1.4133 \pm 0.143$ & $-1.0997 \pm 0.269$ & $0.3822 \pm 1.634$ & $-0.6498 \pm 0.501$ & $-0.9304 \pm 0.335$ & $-0.8904 \pm 1.092$ & $-0.6628 \pm 0.551$ & $-0.7459 \pm 0.880$ & $-0.4793 \pm 0.952$ & $-0.0571 \pm 1.053$ & $-0.8776 \pm 1.354$ \\
Charges Refused & $-1.3137 \pm 0.279$ & $-1.0539 \pm 0.376$ & $-0.9652 \pm 0.346$ & $-1.1835 \pm 0.419$ & $-0.9499 \pm 0.366$ & $-0.9317 \pm 0.245$ & $-0.5424 \pm 0.375$ & $1.8592 \pm 0.595$ & $-1.4997 \pm 0.408$ & $-0.7347 \pm 0.221$ & $-0.6632 \pm 0.167$ \\
Probable Cause Found & $0.0238 \pm \mathrm{NA}$ & $-0.5584 \pm \mathrm{NA}$ & $0.7285 \pm \mathrm{NA}$ & $-0.2287 \pm \mathrm{NA}$ & $-0.7863 \pm \mathrm{NA}$ & $-0.7144 \pm \mathrm{NA}$ & $1.0586 \pm \mathrm{NA}$ & $-1.8991 \pm \mathrm{NA}$ & $-0.7826 \pm \mathrm{NA}$ & $-1.0201 \pm \mathrm{NA}$ & $-0.8464 \pm \mathrm{NA}$ \\
Pretrial Conference Held & $1.0562 \pm 0.218$ & $-0.5267 \pm 0.163$ & $-0.1514 \pm 0.220$ & $0.3589 \pm 0.429$ & $-0.2009 \pm 0.271$ & $-0.4170 \pm 0.403$ & $-0.6075 \pm 0.099$ & $-0.6226 \pm 0.228$ & $-0.0702 \pm 0.746$ & $-0.5237 \pm 0.238$ & $0.2666 \pm 0.546$ \\
Pled Not Guilty & $-1.1237 \pm 1.161$ & $-1.3613 \pm 0.423$ & $0.3605 \pm 1.817$ & $0.2074 \pm 1.131$ & $-0.2693 \pm 0.772$ & $-1.6533 \pm 0.602$ & $-0.3511 \pm 0.583$ & $-1.0901 \pm 0.030$ & $-0.0389 \pm 1.463$ & $-0.1297 \pm 0.734$ & $0.1869 \pm 1.057$ \\
Continuance Occurred & $-0.4406 \pm 0.486$ & $-0.7398 \pm 0.215$ & $-0.4153 \pm 1.205$ & $-0.4957 \pm 0.534$ & $-1.0767 \pm 0.248$ & $-0.9327 \pm 0.196$ & $-1.1960 \pm 0.261$ & $-0.6100 \pm 0.368$ & $-0.2894 \pm 0.300$ & $-0.6624 \pm 0.444$ & $-0.7303 \pm 0.488$ \\
In Custody, Not in Court & $-0.5957 \pm 0.521$ & $-0.5703 \pm 0.790$ & $-0.2778 \pm 2.317$ & $-0.8489 \pm 0.322$ & $-0.6579 \pm 0.238$ & $-0.1491 \pm 0.643$ & $-1.1476 \pm 0.050$ & $-0.5190 \pm 0.846$ & $-0.6398 \pm 1.267$ & $-0.8642 \pm 0.369$ & $-0.9662 \pm 0.639$ \\
Pled Guilty & $1.0953 \pm 0.202$ & $1.5729 \pm 0.129$ & $0.7298 \pm 0.370$ & $0.8908 \pm 0.409$ & $1.4114 \pm 0.491$ & $1.4482 \pm 0.046$ & $0.4924 \pm 0.245$ & $0.4578 \pm 0.450$ & $1.3351 \pm 0.429$ & $1.1585 \pm 0.953$ & $1.3516 \pm 0.225$ \\
Trial Verdict Reached & $0.2156 \pm 0.246$ & $-0.2437 \pm 0.305$ & $-0.1286 \pm 0.745$ & $-0.5678 \pm 0.471$ & $-0.4323 \pm 0.264$ & $-0.0942 \pm 0.391$ & $-0.2558 \pm 0.369$ & $-0.6312 \pm 0.522$ & $-0.3551 \pm 0.423$ & $0.4306 \pm 0.898$ & $-0.0335 \pm 0.682$ \\
Sentence Includes Jail Time & $0.8186 \pm 0.152$ & $0.9018 \pm 0.311$ & $1.0703 \pm 0.592$ & $0.3906 \pm 0.327$ & $0.5738 \pm 0.404$ & $1.1345 \pm 0.484$ & $-0.2189 \pm 0.210$ & $0.3101 \pm 0.569$ & $0.8562 \pm 0.449$ & $0.8443 \pm 0.792$ & $0.9884 \pm 0.433$ \\
Sentence Excludes Jail Time & $0.9747 \pm 0.327$ & $1.3392 \pm 0.211$ & $-0.0726 \pm 0.173$ & $0.2350 \pm 0.351$ & $1.3493 \pm 0.404$ & $0.9556 \pm 0.446$ & $1.8009 \pm 0.089$ & $0.1891 \pm 0.793$ & $0.8028 \pm 0.620$ & $0.5995 \pm 0.770$ & $1.2034 \pm 0.498$ \\
\bottomrule
\end{tabular}%
}
\end{table*}

\section{Implementation Details}
\label{sec:implementation}

The complete experimental pipeline, including BERT embedding pre-computation, 5-fold cross-validation training of the transformer, the random and MLP baselines, all ablation runs, and the interpretability analyses, took approximately 11 hours.

\paragraph{Transformer Architecture}
This model uses a hidden dimension of 896, 4 transformer encoder layers, 8 attention heads, a feed-forward dimension of 2048, GELU activation functions, and dropout of 0.2. The maximum sequence length is 200. Learned positional embeddings are initialized from $\mathcal{N}(0, 0.02^2)$ and are added to each step. A causal mask is applied at every layer so that proceeding $t$ attends to only the proceedings $\le t$. The fuse layer that maps the concatenated proceeding representation to the model hidden dimension is represented as Linear $\rightarrow$ LayerNorm $\rightarrow$ ReLU $\rightarrow$ Dropout. The model has 29,570,766 trainable parameters (approximately 113 MB at 32-bit precision), with the majority concentrated in the Transformer encoder (27.6M parameters, $93\%$ of the total) and the fuse layer (1.4M parameters, $5\%$). All other components, including the defendant, case, Court A section, and prior-label embeddings, together account for less than $2\%$ of the parameter count. Note that the cached BERT text embedding is loaded from disk and contributes no trainable parameters to this model. All training was performed on a single NVIDIA RTX A5000 GPU.

\paragraph{Component Dimensions}
Before fusion, each proceeding representation concatenates: defendant embedding (32), case (charge) embedding (192), BERT text embedding (768), Court A section embedding (32), prior-label embedding (96, from $\mathrm{Linear}(20 \rightarrow 96)$), prior-charge embedding (192, separate linear projection sharing structure with the case embedding), court-type indicator (1), original-bond one-hot (9), cumulative-bond one-hot (9), prior-case-count scalar (1), and a one-hot month-year vector. The month-year dimension is not fixed. Rather, it is set to the number of years observed in training $\times$ 12.

\paragraph{BERT Encoder}
We use \texttt{bert-base-uncased} to encode clerk notes, and its weights are frozen. Text is tokenized, truncated to 512 tokens, and then we extract the final layer CLS representation. Embeddings are pre-computed once and cached to disk. BERT is not fine-tuned at any stage. We selected \texttt{bert-base-uncased} for its simplicity and scalability, and we leave comparison against larger embedding models and against fine-tuned models to future work. 

\paragraph{Optimization}
The model is trained with AdamW (learning rate $2 \times 10^{-5}$, weight decay 0.01), batch size 16, for 20 epochs, with a OneCycleLR schedule (max learning rate $5 \times$ base, 30\% warmup, cosine annealing, initial divisor 25, final divisor $10^4$). Gradients are clipped to norm 1.0, mixed-precision training is enabled, and the loss is unweighted $\mathrm{BCEWithLogitsLoss}$ applied per-label. 

\paragraph{MLP Baseline}
The completion-prediction baseline is a 3-layer feed-forward network: $\mathrm{Linear}(d_{\text{in}} \rightarrow 64) \rightarrow \mathrm{ReLU} \rightarrow \mathrm{Linear}(64 \rightarrow 64) \rightarrow \mathrm{ReLU} \rightarrow \mathrm{Linear}(64 \rightarrow 5)$. Inputs are the features available at the initial proceeding (defendant demographics, initial charges, Court A section), with \texttt{StandardScaler} applied to continuous features and \texttt{LabelEncoder} applied to the Court A section. The model is trained with Adam (learning rate $10^{-3}$), cross-entropy loss over the 5 completion classes (single-label), batch size 16, and 20 epochs. At inference, predictions are taken as the $\mathrm{argmax}$ over the 5 classes.

\paragraph{GRU Baseline}
This baseline replaces the transformer encoder with a two-layer unidirectional GRU whose hidden dimension of 896 matches that of the transformer. A $\mathrm{Linear}(896 \rightarrow 20)$ classification layer is applied to the hidden state at each step, and there is a dropout of 0.2 between GRU layers. The baseline reuses the same proceeding representation $p_t^{(n)}$ as the transformer, including the embedding functions and fusion layer, which are jointly trained with the GRU. The transformer's encoder and positional embeddings are unused. The model is trained with AdamW (learning rate $2 \times 10^{-5}$, weight decay 0.01), batch size 16, for 20 epochs, OneCycleLR scheduling, gradient clipping at norm 1.0, mixed precision, and $\mathrm{BCEWithLogitsLoss}$ applied per-label, which is identical to the transformer.

\section{Data and Code Availability}
\label{sec:dataandcodeavail}

We release the full pipeline (label extraction, feature construction, model training, and all interpretability analyses) at \url{https://github.com/tapilab/docket_sequence_modeling}. We do not release the underlying court records. Expungement in the jurisdiction studied removes a record from public access without destroying it. A static release cannot honor expungement orders issued after publication, and would permanently reintroduce records that defendants are entitled to have removed. Additionally, the clerk notes name victims, witnesses, and other third parties who are not case participants, and hand-entered free text of this kind cannot be reliably de-identified at scale. In place of the source corpus, we release a synthetic dataset matching the schema, field types, and label vocabulary of the real data. 

\section{Macro and Weighted F1 Scores}
\label{sec:macro_weighted}

Tables~\ref{tab:model_performance_macro_weighted} and~\ref{tab:completion_only_performance_macro_weighted} report macro-averaged and weighted-averaged F1 scores for the step-level and completion level evaluations. These complement the micro-averaged results in Tables~\ref{tab:model_performance} and~\ref{tab:completion_only_performance} in the main paper.

\begin{table}[t]
\centering
\footnotesize
\setlength{\tabcolsep}{2pt}
\caption{Step-level performance (mean $\pm$ std)}
\label{tab:model_performance_macro_weighted}
\begin{tabularx}{\columnwidth}{@{}lCC@{}}
\toprule
Model & Macro F1 & Weighted F1\\
\midrule
Transformer
& $.4745 \pm .009$
& $.5671 \pm .012$ \\
\addlinespace
GRU Baseline
& $.4586 \pm .010$
& $.5732 \pm .013$ \\
\addlinespace
Random Baseline
& $.1183 \pm .002$
& $.1824 \pm .004$ \\
\bottomrule
\end{tabularx}
\end{table}

\begin{table}[t]
\centering
\footnotesize
\setlength{\tabcolsep}{2pt}
\caption{Completion-level performance (mean $\pm$ std). Not comparable to Table~\ref{tab:model_performance_macro_weighted}.}
\label{tab:completion_only_performance_macro_weighted}
\begin{tabularx}{\columnwidth}{@{}>{\raggedright\arraybackslash}XCC@{}}
\toprule
Model & Macro F1 & Weighted F1 \\
\midrule
MLP Baseline
& $.2569 \pm .015$
& $.4389 \pm .025$\\
Transformer
& $.4260 \pm .026$
& $.5650 \pm .015$\\
\bottomrule
\end{tabularx}
\end{table}

\section{Threshold-Based Results}
\label{sec:threshold}

The step-level results in the paper select the true number of labels $k_t$ at each proceeding. In this appendix section, we report the performance using labels selected by thresholding the sigmoid output for each label. The transformer model achieves a micro-F1 of $.5213$  using the default $0.5$ binary classification threshold and $.5288$ using a threshold tuned on held-out data within each training fold. These scores are modestly lower than the $.5715$ achieved under top-$k_t$ selection, though they remain in a practically similar range.

\end{document}